\PassOptionsToPackage{table,xcdraw}{xcolor}

\documentclass{article} %
\usepackage{iclr2027_conference,times}

\usepackage[utf8]{inputenc}
\usepackage{graphicx}
\usepackage{wrapfig}
\usepackage{needspace}
\usepackage{booktabs}
\usepackage{amsfonts}
\usepackage{nicefrac}
\usepackage{microtype}
\usepackage{float}
\usepackage{multirow}
\usepackage{tabularx}
\usepackage{array}
\usepackage{pifont}
\usepackage{threeparttable}
\usepackage{longtable}
\usepackage{subcaption}
\usepackage{amsmath}
\usepackage{amsthm}
\theoremstyle{definition}

\theoremstyle{plain}
\newtheorem{proposition}{Proposition}[section]
\theoremstyle{definition}
\graphicspath{{./figures/}}

\renewcommand{\cite}[1]{\citep{#1}}

\newcommand{\RR}{\mathbb{R}}

\usepackage{hyperref}
\usepackage{orcidlink}

\newcommand{\cmark}{\ding{51}}
\newcommand{\xmark}{\ding{55}}
\newcolumntype{L}[1]{>{\raggedright\arraybackslash}m{#1}}
\newcolumntype{C}[1]{>{\centering\arraybackslash}m{#1}}

\author{%
\small
\begin{tabular}{@{}l@{\hspace{1em}}l@{\hspace{1em}}l@{\hspace{1em}}l@{}}
Liheng Ma\thanks{Equal contribution; names are listed alphabetically.} \ $^{1,2}$ & Rui Heng Yang$^{*1}$ & Amin Soleimani Abyaneh\thanks{Work done as interns at Huawei Noah's Ark Lab Montreal} \ $^{2}$ & George Z. Xue$^{\dagger \hspace{0.1em} 3}$ \\
Behnam Rahmati$^1$ & Mateo Cl\'emente$^1$ & Ziwen Hu$^4$ & Anlin Chen$^5$ \\
Tongtong Cao$^5$ & Zhanguang Zhang$^1$\thanks{Correspondenc to: $\langle$\url{zhanguang.zhang@huawei.com}$\rangle$} & Yingxue Zhang$^1$ &
\end{tabular}\\[-0.em]
\footnotesize $^1$  Huawei Noah's Ark Lab Montreal \quad $^2$ McGill University \& Mila-Quebec AI Institute \\ $^3$ University of Toronto \quad $^4$ Huawei Celia Team \quad $^5$ Dep. Foundation Model, Huawei 2012 Labs
}

\iclrfinalcopy %

\begin{document}

\title{Faster-WAM: Do World Action Models Need  Deep Action Modules?}

\maketitle
\suppressfloats[t]

\begin{abstract}
World Action Models (WAMs) build on pretrained video models, whose representations are grounded in physical dynamics and provide a natural basis for action prediction.
Despite this natural foundation, many WAMs still rely on deep, parameter-heavy action-prediction modules that incur high inference latency and may overfit to limited robot demonstrations, restricting their real-world applicability.
In this paper, we advocate a world-model-centric principle that concentrates capacity and computation in the video world model, while a lightweight action expert translates the backbone's representations into executable robot actions.
We realize this principle through three key choices: Dock of Transformers (DoT) with Lite KV-Fusion to give the shallow, lightweight action expert access to representations from all video layers; world-model-only conditioning of the action expert; and retracted 1D-RoPE for 
positional alignment between video keys and action queries.
We test this principle using Faster-WAM, a world-model-centric WAM with only a single-layer action expert.
Despite this restriction on action-specific computation, Faster-WAM achieves competitive control performance on LIBERO and RoboTwin~2.0 without additional embodied pretraining.
It provides approximately \(3.7\times\) and \(1.3\times\) inference speedups over Fast-WAM and \(\pi_{0.5}\), respectively.
Consistent with its world-model-centric design, Faster-WAM demonstrates stronger generalizability under distribution shifts: the same LIBERO-trained policy achieves \(78.3\%\) success on LIBERO-Plus, exceeding Fast-WAM and LingBot-VA by \(26.8\) and \(8.8\) percentage points, respectively. Finally, real-robot experiments demonstrate success rates comparable to Fast-WAM, with substantially lower inference latency and shorter task-completion times.
\end{abstract}

\section{Introduction}

\begin{figure}[t]
  \centering
  \captionsetup{font=small,skip=4pt}
  \captionsetup[subfigure]{font=small,skip=3pt}
  \begin{subfigure}[t]{0.500\linewidth}
    \centering
    \includegraphics[width=\linewidth]{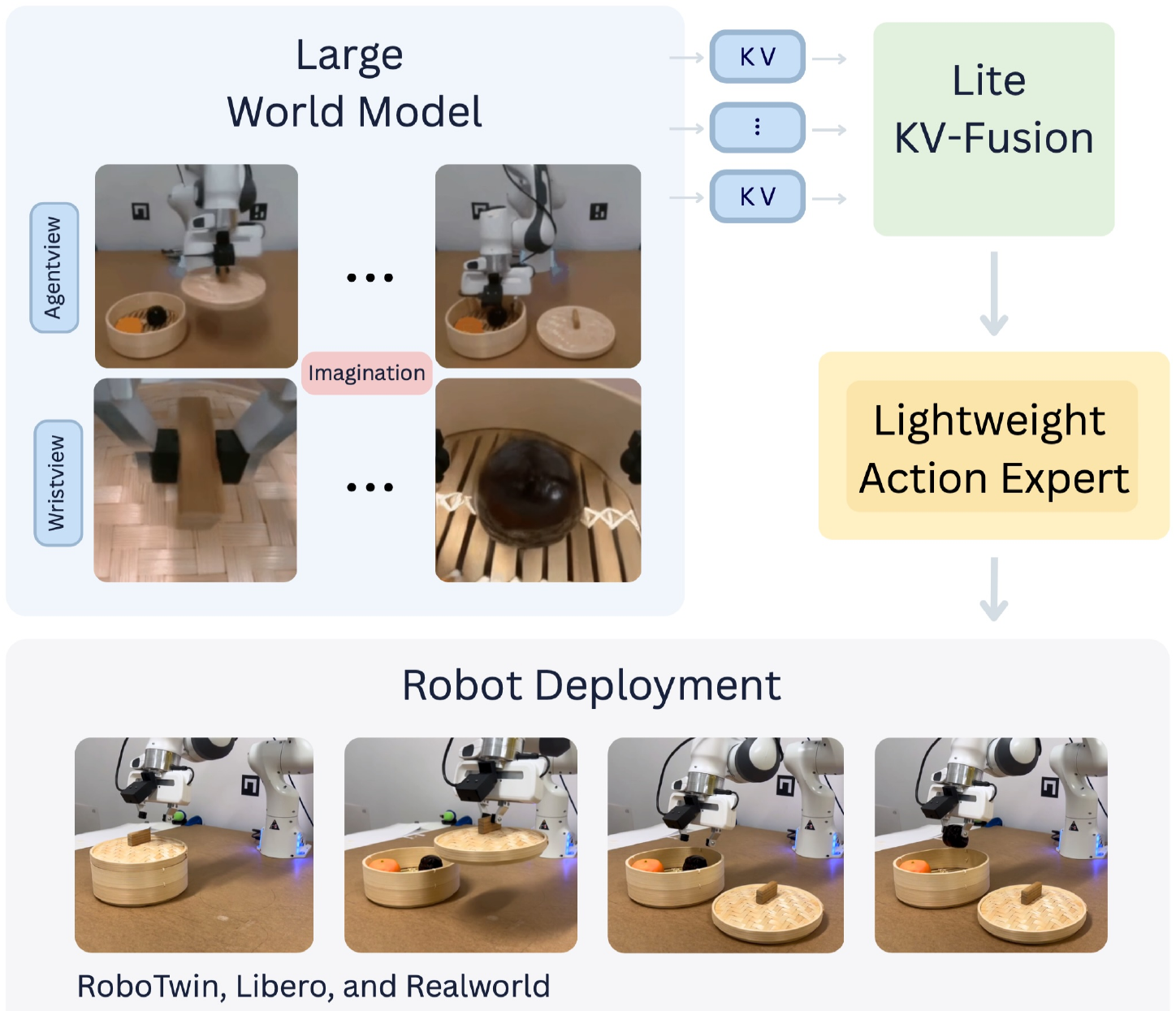}
    \phantomcaption
    \label{fig:firstpage_overview}
  \end{subfigure}\hfill
  \begin{subfigure}[t]{0.480\linewidth}
    \centering
    \includegraphics[width=\linewidth]{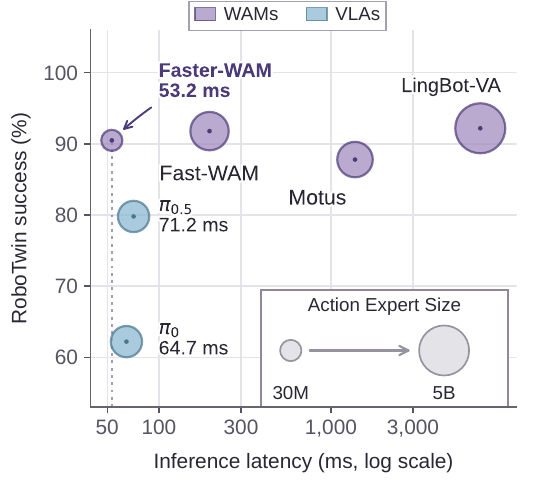}
    \phantomcaption
    \label{fig:robotwin_latency}
  \end{subfigure}
  \caption{(a) Faster-WAM (left) connects a pretrained video world model to a single-layer action head through Lite KV-Fusion, which learns softmax-weighted mixtures of key--value representations across all backbone layers. The fused representations condition action generation for robot control.
  (b) Circle size denotes decoder parameters (compressed scale); LingBot-VA counts its shared 5B backbone.
  Latency measurements for the RoboTwin configuration are reported in Tab.~\ref{tab:robotwin_results}.}
\vspace{-0.5cm}
\end{figure}

World Action Models (WAMs) and Vision-Language-Action (VLA) models are two major paradigms for bridging pretrained foundation models and robot control.
VLAs typically adapt pretrained vision-language models for control~\cite{black2025pi0,black2025pi05}, whereas WAMs build on pretrained video models~\cite{kim2026cosmos}.
By modeling how objects move and interact across space and time, video models learn representations grounded in physical dynamics, providing a natural basis for action prediction.

Despite their natural foundation, many WAMs still devote substantial computation and capacity to action prediction.
Shared-backbone architectures use the large video backbone for action prediction~\cite{kim2026cosmos,li2026causal}\footnote{We classify LingBot-VA as a shared-backbone architecture based on its released implementation, in which video and action tokens traverse the same transformer blocks.}, whereas Mixture-of-Transformers (MoT) architectures introduce separate, slimmer action experts to reduce action-specific computation and capacity~\cite{bi2026motus,yuan2026fastwam}.
In both architectures, however, action prediction still relies on deep, parameter-heavy modules. This raises two concerns.
First, repeated action denoising through large modules incurs high inference latency (Fig.~\ref{fig:robotwin_latency}), thereby limiting real-world applicability (Appx.~\ref{app:real_world_rollout_comparison}).
Second, robot demonstrations are relatively limited compared with video-pretraining corpora.
These high-capacity action-prediction modules may therefore overfit to the limited robot demonstrations, weakening the generalizability afforded by large-scale video pretraining.
These observations raise a central question: \textit{can the pretrained video world model support effective control with substantially less action-specific computation?}

To answer this question, we present \textbf{Faster-WAM}, a \textit{world-model-centric} WAM.
The model concentrates capacity and computation for physical understanding and reasoning in its video backbone.
A lightweight action expert then translates the backbone's representations into robot actions.
We realize this design through three key choices: \\
\textbf{(1) Dock of Transformers (DoT) architecture.}
The proposed DoT architecture connects a lightweight action expert to the video backbone through \textit{Lite KV-Fusion}, which learns, for each action layer, to fuse the native key--value (KV) pairs cached across all video layers.
This decouples the action expert's depth from its access to video-layer representations, which are coupled in MoT architectures where each action layer can only access its corresponding video layer~\cite{yuan2026fastwam,ma2026dit4dit}.
Learning the fusion removes the need to manually select extraction layers, and reusing the native KV cache avoids the additional dense KV reprojection used in prior works.
\textbf{(2) World-model-only conditioning.}
Following the world-model-centric principle, we condition the action expert solely on the video backbone's representations, rather than also providing direct language and proprioceptive inputs as in some prior WAMs~\cite{yuan2026fastwam,cai2026ahawam}.
Routing these inputs only through the world model may reduce over-reliance on them and overfitting to limited robot demonstrations, helping preserve the generalizability afforded by large-scale video pretraining. \\
\textbf{(3) RoPE axis-frequency alignment.}
Because Lite KV-Fusion reuses native video keys, we introduce \textit{retracted 1D-RoPE} to align their rotary positional encodings~\cite{su2024roformer} with those of the action queries.
It removes the fused video keys' original 3D spatiotemporal rotation and re-encodes them using the action queries' 1D temporal axis and frequency grid.
This addresses axis and frequency mismatches in some prior WAMs, which combine action queries with video-derived keys in joint attention without explicit rotary realignment~\cite{yuan2026fastwam,ye2026gigaworldpolicy,cai2026ahawam}. \\
Together, these design choices instantiate the world-model-centric principle in Faster-WAM.

Our experiments further support this world-model-centric design.
Even without additional embodied pretraining and with a single-layer action expert, Faster-WAM achieves competitive control performance on LIBERO~\cite{liu2023libero} and RoboTwin~2.0~\cite{chen2025robotwin}.
It provides approximately \(3.7\times\) and \(1.3\times\) inference speedups over Fast-WAM~\cite{yuan2026fastwam} and \(\pi_{0.5}\)~\cite{black2025pi05}, respectively.
To benchmark generalizability under distribution shifts, we evaluate the same LIBERO-trained policy on LIBERO-Plus~\cite{fei2026liberoplus}, where it outperforms Fast-WAM by \(26.8\) percentage points.
In real-robot experiments, Faster-WAM achieves success rates comparable to those of existing WAMs, with shorter task-completion times. Finally, our ablation studies trace the progression toward Faster-WAM and compare alternative fusion and RoPE designs (Fig.~\ref{fig:design_summary}).

\section{Method}
\label{sec:method}
\label{sec:fasterwam}

\begin{figure}[t]
    \centering
    \includegraphics[width=\linewidth]{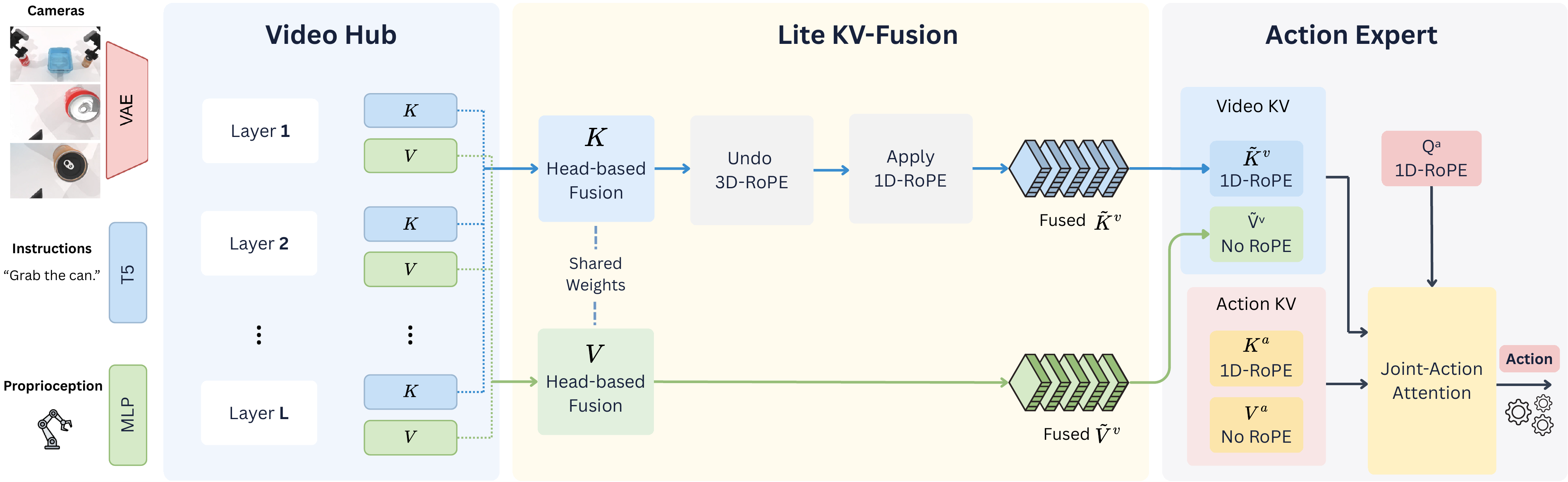}
    \caption{Faster-WAM.
    Lite KV-Fusion connects a shallow action expert to representations across the full video backbone, forming the Dock of Transformers architecture.
    Language and proprioception condition the video backbone, while the action expert receives these inputs through the fused video representations.
    Retracted 1D-RoPE aligns the fused video keys with the action queries before joint attention.}
    \label{fig:kv_fuse}
\end{figure}

At each replanning step, Faster-WAM predicts an action chunk---a sequence of future robot actions---from camera observations, a language instruction, and the robot's proprioceptive state.
It follows a \emph{world-model-centric} design: the pretrained video backbone provides the main capacity for physical understanding and reasoning, while a lightweight action expert translates its representations into robot actions.
Specifically, we use the pretrained Wan2.2-5B video DiT~\cite{wan2025wan} with a single-layer action DiT.
Figure~\ref{fig:kv_fuse} illustrates the three design choices described below: \textbf{Dock of Transformers}, \textbf{world-model-only conditioning}, and \textbf{RoPE axis--frequency alignment}.

\subsection{Dock of Transformers (DoT)}
\label{sec:dot}
\label{sec:kv_fusion}

Video pretraining provides representations grounded in physical dynamics, offering a natural basis for action prediction~\cite{ye2026world}.
To effectively use these representations with a shallow, lightweight action expert, we need an interface that decouples the action expert's depth from its access to video-layer representations.
However, MoT architectures typically couple these two aspects, so reducing action-expert depth can restrict access to some video layers.
We therefore introduce \textbf{Lite KV-Fusion}, an interface within DoT that learns, for each action layer, to fuse the native key--value (KV) pairs cached across all video layers.

\paragraph{Lite KV-Fusion.}
Let \(L_v\) and \(L_a\) denote the depths of the video backbone and action expert, respectively.
Both modules use \(H\) attention heads of dimension \(D\), enabling direct KV reuse.
For video layer \(\ell\) and head \(h\), let \(K^v_{\ell h},V^v_{\ell h}\in\RR^{S_v\times D}\) denote the native cached keys and values for \(S_v\) video tokens.
The cached keys have already undergone the backbone's key normalization and 3D RoPE.

Each action layer \(j\) learns a separate distribution over video layers for each attention head:
\begin{equation}
\begin{aligned}
    \beta_{jh\ell}
    =
    \frac{\exp(\alpha_{jh\ell})}
    {\sum_{r=1}^{L_v}\exp(\alpha_{jhr})}, \qquad
    \widehat K^v_{jh}=\sum_{\ell=1}^{L_v}\beta_{jh\ell}K^v_{\ell h},
    \qquad
    \widehat V^v_{jh}
    =\sum_{\ell=1}^{L_v}\beta_{jh\ell}V^v_{\ell h}.
\end{aligned}
\label{eq:dot_docking}
\end{equation}
\textcolor{black}{%
Here, \(\alpha_{jh\ell}\) are learnable logits, and \(\beta_{jh\ell}\) is the normalized weight assigned to video layer \(\ell\) by head \(h\) of action layer \(j\).}
Since keys and values are strictly paired in Transformers, we fuse them using the same mixing weights.
We initialize all logits \(\alpha_{jh\ell}\) to zero, giving each video layer equal weight, and optimize them jointly with the policy.
Learning the fusion removes the need to manually select extraction layers~\cite{ma2026dit4dit} while giving the shallow, lightweight action expert access to representations from all video layers.

By reusing native KV pairs, Lite KV-Fusion avoids the additional dense projections used by interfaces that construct keys and values from backbone hidden representations~\cite{li2025eagle3,chen2026dflash}.
It adds \(HL_v\) trainable parameters per action layer.
Since Faster-WAM uses a single-layer action expert, Lite KV-Fusion introduces only \(720\) trainable parameters in total.
The fusion costs \(\mathcal O(L_vHD)\) per token, compared with \(\mathcal O(H^2D^2)\) for a dense projection between \(HD\)-dimensional representations.

\paragraph{Joint action attention.}
The action expert integrates the fused video representations through joint attention.
Using normalized and positionally encoded action queries and keys, we compute
\begin{equation}
\begin{aligned}
    A_{jh}
    =
    \operatorname{Softmax}\!\left(
    Q^a_{jh}[K^a_{jh};\widehat{K}^{v}_{jh}]^\top/\sqrt{D}
    +M
    \right), \qquad O^a_{jh}
    =A_{jh}[V^a_{jh};\widehat V^v_{jh}],
\end{aligned}
\label{eq:dot_mixed_attention}
\end{equation}
where \([\,\cdot\,;\,\cdot\,]\) denotes concatenation along the token dimension and \(M\) is an additive attention mask that controls access to future video tokens when present.
The masking constraints are specified in Sec.~\ref{sec:dot_training_inference}.

Together, the shallow, lightweight action expert and Lite KV-Fusion form the \textbf{Dock of Transformers} (DoT) architecture.
Each action layer can access representations from all video layers, decoupling action-expert depth from access to video-layer representations.
This allows \(L_a\) to be chosen independently of \(L_v\).

\paragraph{Compatibility with QK-Norm and RoPE.}
Lite KV-Fusion operates directly on cached keys after any QK-Norm~\cite{henry2020querykey} and RoPE~\cite{su2024roformer} transformations applied by the video backbone.
QK-Norm stabilizes attention by constraining query and key norms, and Lite KV-Fusion preserves the upper bound on key norms (Prop.~\ref{prop:kv_fusion_norm_bounds}).
Its head-specific scalar weights also commute with the shared rotary transform, preserving RoPE positional encoding (Prop.~\ref{prop:kv_fusion_rope}).

\begin{proposition}[Norm-bound preservation in Lite KV-Fusion]
\label{prop:kv_fusion_norm_bounds}
For each token position and attention head, the Euclidean norm of the fused key does not exceed the largest norm among its input keys.
The proof is provided in Appx.~\ref{app:kv_fusion_norm_bounds}.
\end{proposition}

\begin{proposition}[RoPE preservation in Lite KV-Fusion]
\label{prop:kv_fusion_rope}
If all source layers share the same rotary transform at each token position, fusing their rotated keys is equivalent to applying that transform after fusing their unrotated keys.
The proof is provided in Appx.~\ref{app:kv_fusion_rope}.
\end{proposition}

In particular, Prop.~\ref{prop:kv_fusion_rope} permits the exact removal of the native video rotation after fusion, enabling the positional alignment introduced in Sec.~\ref{sec:action_head_design}.

\subsection{World-Model-Only Conditioning}
\label{sec:world_model_conditioning}

Following the world-model-centric principle, we route all conditioning to the action expert through the video backbone.
Existing WAMs typically condition both modules directly on language tokens and proprioceptive states via cross-attention.
We remove these side inputs and cross-attention from the action expert, forcing it to obtain all visual, linguistic, and proprioceptive context from the video world model.
This design reduces the risk of over-reliance on side inputs and grounds action prediction on representations encoded by the video model, 
which is typically more generalizable owing to the large-scale video pretraining.
It also further lightens the action expert and removes the corresponding cross-attention computation from each action denoising step.

\subsection{RoPE Axis--Frequency Alignment}
\label{sec:action_head_design}

In most MoT-based WAMs, The video backbone uses 3D spatiotemporal RoPE, while the action expert uses 1D temporal RoPE.
However, several works combine action queries with video-derived keys in joint attention without explicit rotary realignment~\cite{yuan2026fastwam,ye2026gigaworldpolicy,cai2026ahawam}.
With the native RoPE encodings, 
action queries and video keys use different coordinate axes and frequency grids.
Even the temporal frequency grids differ: video RoPE assigns only part of each head to time, while action RoPE uses the full head. 

\begin{proposition}[RoPE axis--frequency mismatch]
\label{prop:rope_axis_mismatch}
With native 1D action and 3D video RoPE, action indices enter the relative phases of spatial rotary pairs, while mismatched temporal frequencies introduce dependence on absolute indices.
The proof is provided in Appx.~\ref{app:axes_rope}.
\end{proposition}

Therefore, we introduce \textit{retracted 1D-RoPE} to align the fused video keys with the action queries.
For a video token at \(p=(t,y,x)\), let \(R_h^v(p)\) denote its native 3D rotary transform and \(R_h^a(t)\) the action head's 1D transform, with \(t\) expressed on the action temporal scale.
For each action layer \(j\), we compute
\begin{equation}
    k_{jh}^{v\to a}(p)
    =
    R_h^a(t)\,R_h^v(p)^\top\,\widehat k^v_{jh}(p),
    \label{eq:dot_retracted_rope}
\end{equation}
before the action joint attention.
This removes the original 3D rotation via inverse rotation and re-encodes the key using the action queries' temporal axis and frequency grid.
The resulting keys will be used instead of \(\widehat{K}^{v}_{jh}\) in Eq.~\eqref{eq:dot_mixed_attention}; the fused values remain unchanged.
Appx.~\ref{app:rope_derotation} provides the derivation.

In Faster-WAM, action tokens attend only to the observed frame at temporal position \(t=0\), during both training and inference.
Since \(R_h^a(0)=I\), reapplication reduces to the identity.
The video backbone retains its native 3D RoPE whose content can still encode spatial information; retraction removes the explicit rotary factor only from the fused keys supplied to the action expert.

\subsection{Training and Inference}
\label{sec:dot_training_inference}

During training, we jointly optimize the video backbone, action expert, and Lite KV-Fusion.
The video backbone receives gradients from both video and action flow-matching losses, while the action expert receives gradients only from the action loss.
The video backbone does not attend to action tokens, while the action expert jointly attends to action tokens and fused video representations.
Attention masks prevent observed-frame and action tokens from attending to future video tokens.
At inference, we prefill the video backbone's KV cache once per action chunk from the current observation.
Lite KV-Fusion aggregates this cache into a fixed context that the action expert reuses across denoising steps.
Further training and inference details are provided in Appx.~\ref{app:training_details} and ~\ref{app:latency_reproducibility}.

\subsection{Relationship to Related Work}
\label{sec:concurrent_work}

Several recent and concurrent works aim to reduce inference latency of WAMs.
\begin{list}{\textbullet}{%
    \setlength{\leftmargin}{10pt}%
    \setlength{\itemindent}{0pt}%
    \setlength{\labelwidth}{1em}%
    \setlength{\labelsep}{0.5em}%
    \setlength{\itemsep}{2pt}%
    \setlength{\parsep}{0pt}%
    \setlength{\parskip}{0pt}%
    \setlength{\topsep}{0pt}%
    \setlength{\partopsep}{0pt}%
}
    \item \textit{Fast-WAM}~\cite{yuan2026fastwam} eliminates future-video generation by prefilling the observed frame. Faster-WAM inherits this scheme and further reduces action-side computation, achieving a \(3.7\text{--}3.8\times\) speedup over Fast-WAM.
    \item \textit{Light-WAM}~\citep{li2026lightwam} pairs a smaller video backbone with single-pass action prediction. It reports \(76.4\%\) success on RoboTwin, compared with Fast-WAM's \(91.9\%\), a difference of \(15.5\) percentage points.
    \item \textit{Concurrent work} by~\citet{zhao2026faster} retains future-latent conditioning through a single video-expert pass and reduces repeated computation through sparse video--action interaction and interval KV fusion. 
    Their method retains a deep action expert with sparse access to future-conditioned representations, whereas Faster-WAM uses a single-layer action expert conditioned on observed-frame representations. Relative to Fast-WAM, our method achieves a \(3.7\text{--}3.8\times\) inference speedup, compared with approximately \(1.27\times\) for~\citet{zhao2026faster} under their respective measurement settings.
\end{list}

More comprehensive discussion of related work is provided in Appx.~\ref{app:related_work}.

\section{Experiments}
\label{sec:experiments}

We revisit the central question of this work: \textit{can the pretrained video world model support effective control with substantially less action-specific computation?}

To answer it, we evaluate an extreme case of the proposed design: Faster-WAM, a world-model-centric WAM with only a single-layer action expert.
We focus on two questions:
\begin{list}{\arabic{enumi}.}{%
    \usecounter{enumi}%
    \setlength{\leftmargin}{10pt}%
    \setlength{\itemindent}{0pt}%
    \setlength{\labelwidth}{1em}%
    \setlength{\labelsep}{0.5em}%
    \setlength{\itemsep}{1pt}%
    \setlength{\parsep}{0pt}%
    \setlength{\parskip}{0pt}%
    \setlength{\topsep}{2pt}%
    \setlength{\partopsep}{0pt}%
}
    \item Can Faster-WAM retain competitive control performance while reducing inference latency?
    \item Does its world-model-centric design improve generalizability under distribution shifts?
\end{list}

We evaluate control performance and inference latency on LIBERO and RoboTwin~2.0.
To assess generalizability under distribution shifts, we evaluate the same LIBERO-trained policy on LIBERO-Plus.
We further examine whether these benefits extend to real-robot control and use ablation studies to assess the contributions of the three design choices. Lastly, we evaluate the world-model-centric design with an alternative video backbone in Appx.~\ref{app:ltx_backbone}.

\subsection{Competitive Control with Lower Inference Latency}
\label{sec:perform_latency}
\label{sec:libero_robotwin}

To address the first question,
we evaluate Faster-WAM on LIBERO~\cite{liu2023libero} (Tab.~\ref{tab:libero_results}) and RoboTwin~2.0~\cite{chen2025robotwin} (Tab.~\ref{tab:robotwin_results}), following the established protocols of prior WAMs~\cite{bi2026motus,yuan2026fastwam,li2026causal}.
We report task success rate and model-computation latency per action chunk.
The complete dataset, training, evaluation, and latency protocols are provided in Appx.~\ref{app:experimental_details} and~\ref{app:latency_reproducibility}.

\begin{table}[H]
    \centering
    \caption{
        LIBERO success rates (\%) and inference latency per action chunk.
        Baseline success rates are taken from the respective papers.
        Reported mean $\!\pm\!$ std are over three evaluation runs.
        Latency measurement is taken on a 24\,GiB consumer-level GPU. 
        ET denotes pretraining on robot data. 
    }
    \label{tab:libero_results}

    \resizebox{1.\textwidth}{!}{
    \begin{tabular}{l|c|c|cccc|c|c|c}
        \toprule
        \textbf{Method}
        & \textbf{Type}
        & \shortstack{\textbf{ET}}
        & \textbf{Spatial}
        & \textbf{Object}
        & \textbf{Goal}
        & \textbf{Long}
        & \textbf{Average}
        & \shortstack{\textbf{Latency}\\\textbf{(ms)} $\downarrow$}
        & \shortstack{\textbf{Relative}\\\textbf{latency}} \\
        \midrule

        $\pi_0$~\cite{black2025pi0}
        & VLA & \cmark & 96.8 & 98.8 & 95.8 & 85.2 & 94.1 & 65.0 & \(1.3\times\) \\

        $\pi_{0.5}$~\cite{black2025pi05}
        & VLA & \cmark & 98.8 & 98.2 & 98.0 & 92.4 & 96.9 & 67.7 & \(1.3\times\) \\

        X-VLA~\cite{zheng2026xvla}
        & VLA & \cmark & 98.2 & 98.6 & 97.8 & 97.6 & 98.1 & 94.1 & \(1.8\times\) \\

        \noalign{\vskip\aboverulesep}
        \multispan{10}\leaders\hbox{\kern1pt\rule{0.5pt}{0.5pt}\kern1pt}\hfill\cr
        \noalign{\vskip\belowrulesep}

        LingBot-VA~\cite{li2026causal}
        & WAM & \cmark & 98.5 & 99.6 & 97.2 & 98.5 & 98.5
        & 4756.9\textsuperscript{$\ddagger$} & \(91.4\times\)\textsuperscript{$\ddagger$} \\
        Motus~\cite{bi2026motus}
        & WAM & \cmark & 96.8 & 99.8 & 96.6 & 97.6 & 97.7
        & -- & -- \\

        GE-ACT~\cite{liao2026genie}
        & WAM & \cmark & 98.2 & 97.6 & 95.8 & 94.4 & 96.5
        & 200.0\textsuperscript{*} & \(3.8\times\)\textsuperscript{*} \\

        \midrule

        Fast-WAM~\cite{yuan2026fastwam}
        & WAM & \xmark & 98.2 & 100.0 & 97.0 & 95.2 & 97.6
        & 196.0\textsuperscript{$\ddagger$} & \(3.8\times\)\textsuperscript{$\ddagger$} \\

        \textbf{Faster-WAM (Ours)}
        & WAM & \xmark & \(99.1 \pm 0.3\)
        & \(99.5 \pm 0.1\)
        & \(98.4 \pm 0.2\)
        & \(96.2 \pm 0.3\)
        & \(98.3 \pm 0.1\) & 52.0 & \({1.0}\times\) \\

        \bottomrule
    \end{tabular}
    }
  \vspace{-2pt}
            {\scriptsize 
            \begin{flushleft}
            \hspace*{2em}{*} 
            Latency reported in~\citep{liao2026genie}, measured on the same GPU model. \\
            \hspace*{2em}{$\ddagger$}
            Fast-WAM and LingBot-VA latencies combine separately measured components as full pipelines exceed the GPU memory. LingBot-VA timing is based on its released variant.
            \end{flushleft}
            }
\vspace{-2em}
\end{table}
\begin{table}[H]
  \centering
  \caption{RoboTwin 2.0 success rates (\%) and inference latency per action chunk.
  Baseline success rates are taken from the respective papers.
    Reported mean $\!\pm\!$ std are over three evaluation runs.
  ET denotes embodied pretraining on robot data.
  }
  \label{tab:robotwin_results}
  \small
  \resizebox{0.95\linewidth}{!}{%
  \begin{tabular}{l|c|c|cc|c|c|c}
    \toprule
    \textbf{Method} \quad\quad & \textbf{Type} & \quad \textbf{ET} \quad & \quad \textbf{Clean} \quad & \quad \textbf{Rand}. \quad & \quad \textbf{Average} \quad & \shortstack{\textbf{Latency} \\ \textbf{(ms)} $\downarrow$} & \shortstack{\textbf{Relative}\\\textbf{Latency}} \\
    \midrule
    $\pi_0$~\cite{black2025pi0}                         & VLA & \cmark & 65.9 & 58.4 & 62.2 & \(64.7\) & \(1.2\times\) \\
    $\pi_{0.5}$~\cite{black2025pi05}            & VLA & \cmark & 82.7 & 76.8 & 79.8 & \(71.2\) & \(1.3\times\) \\
    \noalign{\vskip\aboverulesep}
    \multispan{8}\leaders\hbox{\kern1pt\rule{0.5pt}{0.5pt}\kern1pt}\hfill\cr
    \noalign{\vskip\belowrulesep}
    Motus~\cite{bi2026motus}                     & WAM & \cmark & 88.7 & 87.0 & 87.8 & \(1380.3\) & \(25.9\times\) \\
    LingBot-VA~\cite{li2026causal}               & WAM & \cmark & 92.9 & 91.5 & 92.2 & \(7386.1\) & \(138.8\times\) \\
    \midrule
    Motus from Wan2.2~\cite{bi2026motus}              & WAM & \xmark  & 77.6 & 77.0 & 77.3 & -- & -- \\
    LingBot-VA from Wan2.2~\cite{li2026causal}         & WAM & \xmark  & 80.6 & --    &  --  & -- & -- \\
    Fast-WAM~\cite{yuan2026fastwam}                 & WAM & \xmark  & 91.9 & 91.8 & 91.8 & \(196.8\) & \(3.7\times\) \\
    \textbf{Faster-WAM (Ours)}                    & WAM & \xmark
    & \(90.4\!\pm\!0.3\)
    & \(90.6\!\pm\!0.2\)
    & \(90.5\!\pm\!0.2\) & \textbf{\(53.2\)} & \({1.0}\times\) \\
    \bottomrule
  \end{tabular}%
  }
\end{table}

\paragraph{Competitive Control Performance} Despite restricting the action expert to a single layer, Faster-WAM achieves \(98.3\pm0.1\%\) average success on LIBERO and \(90.5\pm0.2\%\) on RoboTwin without additional embodied pretraining.
Across both benchmarks, Faster-WAM remains competitive with existing VLAs and WAMs, including those that use additional embodied pretraining.

\paragraph{Faster Inference Speed}
We measure the inference latency of Faster-WAM and the baselines on the same machine with a \(24\,\mathrm{GiB}\) consumer GPU.
Unless otherwise specified, measurements include encoding, conditioning, and action generation.
The complete profiling protocol and results are provided in Appx.~\ref{app:latency_reproducibility}.

Faster-WAM and Fast-WAM use the same action-generation protocol, predicting \(32\)-action chunks with \(10\) denoising steps.
Under this matched setting, Faster-WAM reduces inference latency from \(196.0\) to \(52.0\,\mathrm{ms}\) on LIBERO and from \(196.8\) to \(53.2\,\mathrm{ms}\) on RoboTwin, corresponding to approximately \(3.8\times\) and \(3.7\times\) speedups, respectively.
Faster-WAM also achieves lower inference latency than representative VLAs such as $\pi_0$ and $\pi_{0.5}$ under their native configurations.
The phase-wise breakdown in Appx.~\ref{app:latency_reproducibility} (Fig.~\ref{fig:inference_latency_phases}) shows that the main latency reduction comes from action generation.
Its latency decreases from approximately \(150\,\mathrm{ms}\) in Fast-WAM to \(6\,\mathrm{ms}\) in Faster-WAM, while encoding and conditioning costs remain similar.
Lower inference latency supports responsive real-world control by reducing the delay between observation and action prediction as the robot and its environment evolve.

Together, these results answer the first question: Faster-WAM, with its single-layer action expert, preserves access to the video hierarchy through DoT, yielding competitive control at substantially lower latency.

\subsection{World-Model-Centric Design Improves Generalization}
\label{sec:generalizability}
\label{sec:libero_plus}

To answer our second question, we evaluate Faster-WAM on LIBERO-Plus~\cite{fei2026liberoplus} under an out-of-distribution (OOD) setting.
The model is trained on LIBERO and evaluated across seven perturbation categories in LIBERO-Plus.
Complete evaluation details are provided in Appx.~\ref{app:experimental_libero_evaluation}.

\begin{table}[H]
    \centering
    \caption{
        LIBERO-Plus success rates (\%) under seven perturbation categories.
    Faster-WAM results are reported as mean \(\!\pm\!\) std over three evaluation runs.
    Baseline results are taken from~\citet{zhang2026do,huang2026roboharness}.
    ET denotes embodied pretraining.
    }
    \label{tab:libero_plus_results}

    \setlength{\tabcolsep}{3.5pt}
    \resizebox{\textwidth}{!}{
    \begin{tabular}{l|c|ccccccc|c}
        \toprule
        \textbf{Method}
        & \textbf{ET}
        & \textbf{Background}
        & \textbf{Camera}
        & \textbf{Language}
        & \textbf{Light}
        & \textbf{Object}
        & \textbf{Robot}
        & \textbf{Sensor}
        & \textbf{Overall} \\
        \midrule

        $\pi_0$~\cite{black2025pi0}
        & \cmark & 81.4 & 13.8 & 58.8
        & 85.0 & 68.9 & 6.0 & 79.0 & 53.6 \\

        $\pi_{0.5}$~\cite{black2025pi05}
        & \cmark & 94.6 & \text{75.4} & 85.6
        & \text{96.9} & \text{85.7} & 77.5 & \text{89.7}
        & \text{85.7} \\

        X-VLA~\cite{zheng2026xvla}
        & \cmark & \text{96.0} & 23.4 & 75.7
        & 88.2 & 71.8 & \text{89.7} & 62.7 & 71.4 \\

        LingBot-VA~\cite{li2026causal}
        & \cmark & 53.1 & 40.9 & 86.4
        & 82.3 & 76.2 & 83.0 & 64.4 & 69.5 \\

        GE-ACT~\cite{liao2026genie}
        & \cmark & 86.0 & 60.7 & 77.4
        & 95.8 & 80.2 & 77.0 & 90.9 & 80.3\textsuperscript{*} \\
        
        \midrule

        Fast-WAM~\cite{yuan2026fastwam}
        & \xmark & 53.7 & 16.4 & 68.9
        & 78.2 & 60.7 & 44.5 & 37.7 & 51.5 \\

        \textbf{Faster-WAM (Ours)}
        & \xmark
        & \(60.7 \pm 0.5\)
        & \(76.5 \pm 0.6\)
        & \(90.9 \pm 0.6\)
        & \(94.2 \pm 0.7\)
        & \(81.7 \pm 0.1\)
        & \(58.3 \pm 0.3\)
        & \(84.6 \pm 0.5\)
        & \(78.3 \pm 0.1\) \\

        \bottomrule
    \end{tabular}
    }
    \vspace{-2pt}
        {\scriptsize
       \begin{flushleft}
        \hspace*{2em}       {*}
        GE-ACT uses separate suite-specific checkpoints; Faster-WAM uses one checkpoint shared across all four suites.
       \end{flushleft}
        }
        \vspace{-2pt}
\end{table}

Faster-WAM and Fast-WAM build on the same pretrained video world model, Wan2.2-5B~\cite{wan2025wan}, and perform comparably on standard LIBERO.
On LIBERO-Plus, Faster-WAM achieves \(78.3\%\) overall success, exceeding Fast-WAM by \(26.8\) percentage points and outperforming it across all seven perturbation categories (Tab.~\ref{tab:libero_plus_results}).
Without additional embodied pretraining or data augmentation, Faster-WAM also outperforms \(\pi_0\), X-VLA, and LingBot-VA, trailing only GE-ACT and \(\pi_{0.5}\) among the compared methods.
The GE-ACT result is not directly comparable because it uses separate suite-specific checkpoints, whereas Faster-WAM uses one checkpoint shared across all four suites.
This broad improvement suggests that concentrating capacity in the video backbone better preserves the generalization benefits of the large-scale pretrained video world model.

\subsection{Additional Study}
\subsubsection{Real-World Control with Faster-WAM}
\label{sec:real_world}

We next evaluate whether Faster-WAM's favorable performance--latency trade-off transfers to physical control.
Using a Franka Panda arm, we consider four tasks spanning pick-and-place, occluded sequential manipulation, contact-sensitive pushing, and long-horizon object arrangement.
Each policy is jointly trained on \(100\) demonstrations per task and evaluated over \(20\) rollouts per task, except LingBot-VA on \texttt{dish\_utensil}, for which we use \(10\) rollouts due to its substantially higher inference latency.
Complete experimental details are provided in Appx.~\ref{app:real_robot_setup}.

\begin{table}[H]
    \centering
    \caption{\textbf{Real-world performance on the Franka Panda arm.}
    Average completion time and average inference latency per action chunk are measured over successful rollouts on the deployment workstation.}
    \label{tab:real_world_results}
    \small
    \setlength{\tabcolsep}{5pt}
    \resizebox{1.0\linewidth}{!}{%
    \begin{tabular}{lc*{4}{cc}c|c}
        \toprule
        \multirow{2}{*}{\textbf{Method}}
        & \multirow{2}{*}{\textbf{ET}}
        & \multicolumn{2}{c}{\textbf{place\_cube}}
        & \multicolumn{2}{c}{\textbf{pick\_fruit}}
        & \multicolumn{2}{c}{\textbf{push\_T}}
        & \multicolumn{2}{c}{\textbf{dish\_utensil}}
        & \multirow{2}{*}{\textbf{Avg. SR (\%)} $\uparrow$}
        & \multirow{2}{*}{\textbf{Latency (ms)} $\downarrow$} \\
        \cmidrule(lr){3-4}\cmidrule(lr){5-6}\cmidrule(lr){7-8}\cmidrule(lr){9-10}
        & & SR (\%) $\uparrow$ & CT $\downarrow$ & SR (\%) $\uparrow$ & CT $\downarrow$ & SR (\%) $\uparrow$ & CT $\downarrow$ & SR (\%) $\uparrow$ & CT $\downarrow$ & & \\
        \midrule
        $\pi_{0.5}$ & \cmark & 100 & 10.8 & 100 & 24.4 & 55 & 22.8 & 90 & 39.0 & 86.3 & 151 \\
        LingBot-VA & \cmark & 100 & 117.8 & 90 & 175.3 & 10 & 112.5 & 10 & 269.0 & 52.5 & 10{,}897 \\
        Fast-WAM & \xmark & 95 & 21.9 & 95 & 36.7 & 70 & 24.6 & 60 & 85.0 & 80.0 & 502 \\
        \midrule
        \textbf{Faster-WAM} & \xmark & 100 & 18.2 & 90 & 35.1 & 65 & 23.8 & 70 & 72.0 & {81.3} & 129 \\
        \bottomrule
    \end{tabular}%
    }
    \par\vspace{3pt}
    \begin{minipage}{0.6\linewidth}
        \tiny\centering
        \textbf{SR}: Success rate (\%) \hspace{5pt} \textbf{CT}: Completion time (s) \hspace{5pt} \textbf{ET}: Embodied pretraining
    \end{minipage}
\end{table}

As shown in Tab.~\ref{tab:real_world_results}, without embodied pretraining, Faster-WAM achieves an average success rate of \(81.3\%\), comparable to Fast-WAM's \(80.0\%\).
The embodied-pretrained \(\pi_{0.5}\) attains the highest measured success rate of \(86.3\%\).
LingBot-VA achieves \(52.5\%\) under the same downstream setting: it performs strongly on the two pick-and-place tasks but drops to \(10\%\) on both the contact-sensitive \texttt{push\_T} and long-horizon \texttt{dish\_utensil} tasks.
We follow LingBot-VA's released training code~\cite{li2026causal}.

On the deployment workstation, Faster-WAM reduces inference latency from Fast-WAM's \(502\,\mathrm{ms}\) to \(129\,\mathrm{ms}\) per action chunk, a \(3.9\times\) speedup, and is slightly faster than \(\pi_{0.5}\) at \(151\,\mathrm{ms}\).
LingBot-VA requires \(10{,}897\,\mathrm{ms}\) per chunk, which limits its feasibility for real-world robot tasks requiring frequent replanning.
Although these absolute values are not directly comparable to the controlled profiling results in Sec.~\ref{sec:perform_latency}, the speedup over Fast-WAM remains consistent (\(3.9\times\) versus \(3.8\times\)).

Faster-WAM also achieves lower mean completion times than all evaluated WAM baselines on all four tasks.
Despite its slightly higher per-chunk inference latency, \(\pi_{0.5}\) achieves shorter mean completion times than Faster-WAM on all four tasks.
These results distinguish inference efficiency from overall task execution speed: faster policy calls do not necessarily yield shorter completion times.

Together, these results show that Faster-WAM retains competitive real-world task success with a single-layer action expert while reducing both inference latency and mean completion time relative to the evaluated WAM baselines.
Illustrative real-world rollouts are provided in Appx.~\ref{app:real_world_rollouts}.

\begin{figure}[h]
  \centering
  \captionsetup{font=small,skip=3pt}
  \begin{minipage}[t]{0.45\linewidth}
    \vspace{0pt}
    \centering
    \includegraphics[width=\linewidth]{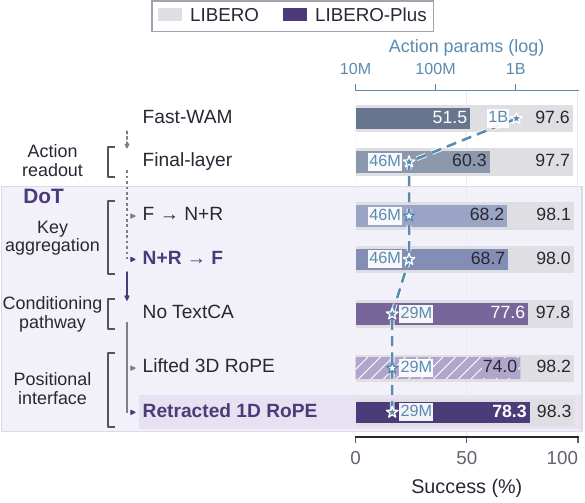}
    \caption{{Steps towards Faster-WAM.}
    Branches show alternatives, boldface marks the adopted choices, and stars indicate action-head parameter counts.
F: fusion; N: QK-Norm; R: RoPE.}
    \label{fig:design_summary}
  \end{minipage}\hfill
  \begin{minipage}[t]{0.49\linewidth}
    \vspace{0pt}
    \centering
    \includegraphics[width=\linewidth]{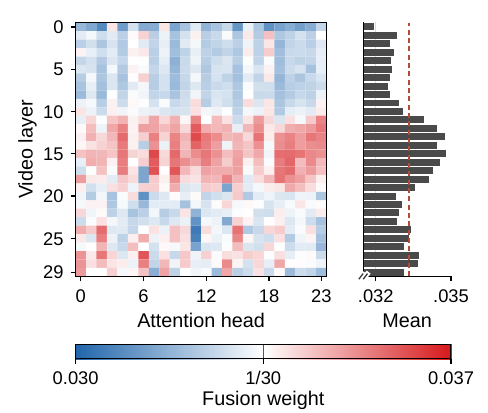}
    \caption{Learned KV-fusion weights by layer and head (left) of Faster-WAM on Libero, with the mean across heads (right).
    White and the dashed line mark \(1/30\).}
    \label{fig:kv_fusion_signal}
    \label{fig:design_ablation}
  \end{minipage}
\end{figure}

\subsubsection{Ablating the World-Model-Centric Design}
\label{sec:ablations}

LIBERO measures in-distribution control, while LIBERO-Plus perturbs the same task suites to probe OOD generalization.
Fig.~\ref{fig:design_summary} traces how each design choice affects these two regimes.
A single-layer action expert already improves OOD generalization over Fast-WAM, even when it accesses only the final video layer, while preserving in-domain control.
Lite KV-Fusion further improves both regimes by exposing the shallow expert to the full hierarchy of world-model representations.
We compare two variants: F\(\rightarrow\)N+R fuses keys before QK-Norm and RoPE and then applies QK-Norm and RoPE after fusion, while N+R\(\rightarrow\)F directly fuses the processed keys.
They perform comparably, with a small OOD advantage for N+R\(\rightarrow\)F.
Because this variant also directly reuses the backbone's native KV cache, we adopt it thereafter.

We next remove action-side text cross-attention (TextCA) and direct proprioceptive conditioning, requiring the action expert to obtain all context through the video world model.
This produces a substantial OOD gain with a small in-domain drop.
We hypothesize that the mismatch between the video and action RoPE bases becomes more consequential once precise proprioceptive information must pass through the video representations.
Both alignment methods recover the in-domain performance, consistent with this explanation.
Lifted 3D RoPE, however, loses part of the OOD gain, possibly because assigning pseudo spatial anchors to action tokens introduces an additional spatial prior.
Retracted 1D-RoPE avoids this assumption and retains stronger OOD generalization; we therefore adopt it in Faster-WAM.

\subsubsection{Visualization of Lite KV-Fusion}
\label{sec:kv_fusion_visualization}

Fig.~\ref{fig:kv_fusion_signal} visualizes the layer weights learned by Lite KV-Fusion on LIBERO.
The learned mixtures are non-uniform and vary across attention heads, with different heads emphasizing different depths of the video backbone.
Together with the improvement over final-layer readout in Fig.~\ref{fig:design_summary}, this suggests that Lite KV-Fusion draws complementary information from the video hierarchy instead of collapsing to a single extraction layer.
It provides this adaptive readout with only \(720\) mixing logits, avoiding manual layer selection as used in DiT4DiT~\cite{ma2026dit4dit}.

\section{Limitations}

Our experiments do not include additional embodied pretraining.
Whether it further improves control performance and generalizability under the world-model-centric design remains to be investigated.
We validate DoT with two video world models, Wan2.2 and LTX-Video.
Its applicability to newer video world models, such as Wan3.0-Video, and to non-generative predictive video models, such as V-JEPA 2~\cite{assran2025vjepa2}, remains to be explored.
We evaluate Faster-WAM as an extreme case with a single-layer action expert, without conducting an exhaustive search for the optimal depth.
Our results support the world-model-centric principle by showing that competitive control does not necessarily require deep, parameter-heavy action experts.
They do not establish that a single layer is optimal or that increasing depth is detrimental.
Further study is needed to understand how action-expert depth affects control performance, generalizability, and inference latency.
Finally, our physical evaluation is limited to one robot platform and four manipulation tasks.
Broader evaluation across robot embodiments, task families, and deployment conditions is needed to assess how well DoT generalizes to other real-world settings.

\section{Conclusion}

We asked whether World Action Models need deep action modules.
Our results suggest that they need not: action-expert depth can be decoupled from access to the video hierarchy, allowing the pretrained world model to provide the primary capacity for physical understanding and reasoning.
Following this world-model-centric principle, Faster-WAM uses a single-layer action expert.
Its DoT architecture employs Lite KV-Fusion to access representations across all video layers, while world-model-only conditioning and retracted 1D-RoPE route and align contextual information through the video backbone.

Without additional embodied pretraining, Faster-WAM retains competitive control performance on LIBERO and RoboTwin~2.0 while reducing matched inference latency by \(3.7\text{--}3.8\times\) over Fast-WAM.
The same LIBERO-trained policy improves success on LIBERO-Plus by \(26.8\) percentage points, and real-robot experiments further show competitive success with lower online latency and shorter completion times.
These findings suggest that making better use of pretrained world-model representations can support efficient and generalizable control with little action-specific computation.

\newpage

\bibliography{references_clean}
\bibliographystyle{iclr2027_conference}

\clearpage
\appendix
\clearpage
\section{Related Work}
\label{app:related_work}
\paragraph{Vision-Language-Action Models (VLAs).}
Collecting diverse robot demonstrations is costly, motivating the transfer of knowledge from large-scale pretraining.
VLAs adapt pretrained vision-language representations for control~\citep{zitkovich2023rt2,kim2025openvla,black2025pi0,black2025pi05,li2024cogact,nvidia2025gr00t,zheng2026xvla,yuan2026qwenrobotmanip}.
A VLA typically builds on a large language model (LLM) backbone with a vision adapter and, in some architectures, an action expert~\cite{black2025pi0,black2025pi05}.
Thus, VLAs typically need to bridge language-oriented representations to continuous robot control.
To equip VLAs with stronger visual perception and visual foresight, subsequent work augments these policies with future-image, visual-subgoal, or latent-state prediction~\citep{cen2025worldvla,zhao2025cotvla,cai2026internvlaa1,zhang2025dreamvla,sun2026vlajepa}.

\paragraph{World Action Models (WAMs).}
WAMs build on pretrained video world models to support the spatial reasoning and dynamics understanding needed for robot control~\citep{du2023learning,bi2026motus,li2026causal,ye2026world}.
Unlike VLAs, 
video world models learn representations grounded in physical dynamics, providing a natural basis for action prediction.

However, the use of large video backbones and iterative video generation makes WAM inference computationally expensive, limiting its applicability to real-time robot control.
Thus, recent work are moving from shared-backbone designs toward Mixture-of-Transformers (MoT) architectures, which decouple video and action generation.
Shared-backbone designs use the full world-model backbone for both video and action denoising~\citep{kim2026cosmos,ye2026world,li2026causal}\footnote{\cite{li2026causal} introduces both shared-backbone and MoT variants, but its publicly released code primarily implements the shared-backbone variant.}.
Although effective, these designs typically incur high inference latency.
MoT architectures decouple the video and action generation by using separate video and action experts
With narrower action modules and distinct generation frequencies and denoising schedules for video and action~\citep{bi2026motus,li2026causal}, 
the inference latency can be significantly improved.
Fast-WAM further shows that observed-frame prefilling is sufficient to provide visual knowledge from the video backbone without explicit future-video denoising~\citep{yuan2026fastwam},
which further reduce the inference latency.

Nonetheless, most MoT-based WAMs, including Fast-WAM, typically couple the action expert's depth with its access to representations across the video backbone.
Consequently, reducing the expert's depth typically restrict its direct access to some backbone layers~\cite{ma2026dit4dit}
Accessing representations from all video layers therefore requires a deep action expert that must run repeatedly during action denoising, leaving WAM inference slower than that of representative VLAs.
Faster-WAM pursues a complementary architectural improvement guided by the world-model-centric principle: DoT decouples action depth from video depth while preserving access to the full video hierarchy.
Faster-WAM thereby directly reduces inference latency while preserving competitive control performance.

\paragraph{Latent Feature and KV Aggregation.}
A shallow action expert also needs an efficient interface to representations across the video backbone.
DiT4DiT~\cite{ma2026dit4dit} and mimic-video~\cite{pai2025mimicvideo} introduce action decoders conditioned on intermediate video hidden states, with extraction layers selected empirically.
In LLM speculative decoding, the EAGLE family reuses target-model features for lightweight draft models, with EAGLE-3 fusing hidden states from multiple depths through a dense projection~\citep{li2024eagle,li2024eagle2,li2025eagle3}.
DFlash~\cite{chen2026dflash} similarly fuses selected hidden states and projects them into a lightweight diffusion drafter as keys and values.
Our Lite KV-Fusion instead learns head-specific mixtures of the video backbone's native keys and values across all layers, avoiding manual layer selection and additional dense projections of backbone features.

\clearpage
\section{Theoretical Support}

\subsection{Norm-Bound Preservation under Lite KV-Fusion}
\label{app:kv_fusion_norm_bounds}

Lite KV-Fusion is a standalone aggregation operator whose mixing weights depend on the output layer and head.
Fix one instance of this operator, a token position \(p\), and a head \(h\).
Let \(\widetilde k_{\ell h}(p)\in\RR^d\) denote the cached key from source layer \(\ell\), after native key normalization and RoPE, when applicable.
The operator produces
\[
    \widehat k_h(p)=\sum_{\ell=1}^{L_v}\beta_{h\ell}\widetilde k_{\ell h}(p),
    \qquad \beta_{h\ell}\geq 0,
    \qquad \sum_{\ell=1}^{L_v}\beta_{h\ell}=1,
\]
where the last two properties follow from softmax normalization over source layers.

\begin{proof}[Proof of Prop.~\ref{prop:kv_fusion_norm_bounds}]
The triangle inequality and nonnegative mixing weights give
\begin{equation}
\begin{aligned}
    \|\widehat k_h(p)\|_2
    &\leq \sum_{\ell=1}^{L_v}\beta_{h\ell}\|\widetilde k_{\ell h}(p)\|_2\\
    &\leq \max_{\ell}\|\widetilde k_{\ell h}(p)\|_2
           \sum_{\ell=1}^{L_v}\beta_{h\ell}\\
    &=\max_{\ell}\|\widetilde k_{\ell h}(p)\|_2.
\end{aligned}
\label{eq:kv_fusion_norm_bound}
\end{equation}
Thus, the fusion operator preserves any common upper bound on its input key norms within a head.
\end{proof}

This result retains scale control on attention keys introduced by QK-Norm.
Note that this does not guarantee the exact equal key norms.

\subsection{RoPE Preservation under Lite KV-Fusion}
\label{app:kv_fusion_rope}

Lite KV-Fusion aggregates keys at the same token position across video layers.
Because these keys share a rotary transform, fusion preserves that transform exactly.

Fix an output layer and attention head \(h\), and let \(k_{\ell h}(p)\in\RR^d\) be the key at position \(p\) in video layer \(\ell\), after key normalization but before RoPE.
Here, \(p\) denotes a temporal or spatiotemporal coordinate.
All video layers use the same rotary matrix \(R_h(p)\), with the same frequencies, axis partition, and channel pairing.
Their cached keys are therefore \(\widetilde k_{\ell h}(p)=R_h(p)k_{\ell h}(p)\).
The softmax layer weights \(\beta_{h\ell}\) are independent of token position and shared across all channels within a head.
Softmax acts only on the layer logits, so fusion remains linear in the keys.
Define
\begin{equation}
\begin{aligned}
    \bar k_h(p)&=\sum_{\ell=1}^{L_v}\beta_{h\ell}k_{\ell h}(p),\\
    \widehat k_h(p)&=\sum_{\ell=1}^{L_v}\beta_{h\ell}\widetilde k_{\ell h}(p).
\end{aligned}
\label{eq:appx_rope_fusion_keys}
\end{equation}

Prop.~\ref{prop:kv_fusion_rope} states that the fused video key retains the shared rotary transform:
\begin{equation}
    \widehat k_h(p)=R_h(p)\bar k_h(p).
    \label{eq:appx_rope_fusion_commutes}
\end{equation}

\begin{proof}[Proof of Prop.~\ref{prop:kv_fusion_rope}]
Each \(\beta_{h\ell}\) is a scalar shared across the channels of head \(h\), and \(R_h(p)\) is identical across video layers at position \(p\).
Therefore,
\[
\begin{aligned}
    \widehat k_h(p)
    &=\sum_{\ell=1}^{L_v}\beta_{h\ell}\widetilde k_{\ell h}(p)\\
    &=\sum_{\ell=1}^{L_v}\beta_{h\ell}R_h(p)k_{\ell h}(p)\\
    &=R_h(p)\sum_{\ell=1}^{L_v}\beta_{h\ell}k_{\ell h}(p)
    =R_h(p)\bar k_h(p).
\end{aligned}
\]
Thus, fusion preserves the source video keys' rotary transform, including its axes, frequencies, and channel pairing.
This proves Eq.~\eqref{eq:appx_rope_fusion_commutes} for both 1D and 3D RoPE.
\end{proof}

This identity permits exact de-rotation after fusion, as detailed in Appx.~\ref{app:rope_derotation}.

\subsection{Misalignment between 1D Temporal and 3D Video RoPE}
\label{app:axes_rope}

Consider one attention head with \(d/2\) rotary pairs, and let \(q_j,k_j\in\RR^2\) denote the query and key in pair \(j\), after any normalization.
For a two-dimensional rotation \(R(\theta)\), RoPE satisfies~\cite{su2024roformer}
\begin{equation}
    (R(\theta)q_j)^\top R(\theta')k_j
    =q_j^\top R(\theta'-\theta)k_j.
    \label{eq:rope_rel}
\end{equation}
The usual relative-position interpretation requires matching coordinate axes and frequencies in corresponding query and key pairs.

For the native frequency schedules, 1D action RoPE assigns all \(d/2\) pairs to time, with
\(\nu_j=\theta^{-2j/d}\), \(j=0,\ldots,d/2-1\), for base \(\theta>1\).
The video backbone partitions the head into even dimensions \(d_t+d_y+d_x=d\), using
\(\omega^b_r=\theta^{-2r/d_b}\) for axis \(b\in\{t,y,x\}\).
Thus, video RoPE has only \(d_t/2\) temporal frequencies, with a different frequency grid even when both encodings use the same base \(\theta\).
For \(d=128\), the video partition is \((d_t,d_y,d_x)=(44,42,42)\): action RoPE uses \(64\) temporal frequencies, compared with \(22\) for video time and \(21\) for each spatial axis.
Both encodings still use \(64\) rotary pairs in total.

\begin{proof}[Proof of Prop.~\ref{prop:rope_axis_mismatch}]
\label{proof:rope_axis_mismatch}
Let \(i\) denote the action index, \(u=(t,y,x)\) the video coordinates, and \(b(j)\in\{t,y,x\}\) the video axis assigned to pair \(j\).
For positive action and video frequencies \(\nu_j,\omega_j\), respectively, Eq.~\eqref{eq:rope_rel} gives the unscaled attention logit
\begin{equation}
    \mathcal S(i,u)
    =\sum_{j=0}^{d/2-1}
      q_j^\top R\!\left(\omega_j u_{b(j)}-\nu_j i\right)k_j.
    \label{eq:appx_axis_mismatch_logit}
\end{equation}
\textcolor{black}{For a spatial pair with coordinate \(z\in\{y,x\}\), its relative phase can be written as}
{\color{black}
\begin{equation}
    \omega_j z-\nu_j i
    =\omega_j\bigl(z-\rho_j(i)\bigr),
    \qquad
    \rho_j(i)=\frac{\nu_j}{\omega_j}i.
    \label{eq:appx_spurious_spatial_offset}
\end{equation}
}
\textcolor{black}{
Thus, the action index acts as an implicit spatial reference.
Advancing it by \(\delta\) shifts this unwrapped reference by \((\nu_j/\omega_j)\delta\), even when the observed frame (\(t=0\)) and its spatial content are fixed.
This artificial displacement is determined by token order and frequency ratios, independently of scene geometry.}

For a temporal pair, the relative phase is
\begin{equation}
    \omega_j t-\nu_j i
    =\omega_j(t-i)+(\omega_j-\nu_j)i.
    \label{eq:appx_temporal_frequency_mismatch}
\end{equation}
When \(\omega_j\ne\nu_j\), this phase depends on the absolute action index as well as the time difference.
Shifting both \(t\) and \(i\) by \(\delta\) changes the phase by \((\omega_j-\nu_j)\delta\), although \(t-i\) is unchanged.
With the native schedules above, \(\omega_j=\theta^{-2j/d_t}\) and \(\nu_j=\theta^{-2j/d}\) on the temporal block, so their frequencies differ for \(j>0\).
Together, the axis assignment and frequency-grid mismatch prevent a consistent relative-position interpretation.
\end{proof}

\subsection{Retracted 1D RoPE for Video Tokens}
\label{app:rope_derotation}

We align fused video keys with the action module's 1D RoPE by removing their native 3D rotation and applying the action rotary basis.

\paragraph{De-rotation.}
RoPE rotates each pair of key channels by an angle determined by the token position.
De-rotation applies the opposite angle to the same pair.
For a video token at \(p=(t,y,x)\), pair \(r\) uses
\(\theta_r(p)=\omega_r p_{b(r)}\), where \(b(r)\in\{t,y,x\}\) selects its axis and \(\omega_r\) is its native RoPE frequency.
Let \(\widehat k_{h,r}(p)\in\RR^2\) denote this pair in the fused key of head \(h\).
The de-rotated pair is
\begin{equation}
\begin{aligned}
    \bar k_{h,r}(p)
    &=R(-\theta_r(p))\widehat k_{h,r}(p)\\
    &=
    \begin{pmatrix}
        \cos\theta_r(p) & \sin\theta_r(p)\\
        -\sin\theta_r(p) & \cos\theta_r(p)
    \end{pmatrix}
    \widehat k_{h,r}(p).
\end{aligned}
\label{eq:appx_rope_derotation_pair}
\end{equation}
Since \(R(-\theta)=R(\theta)^\top=R(\theta)^{-1}\), applying this operation to every pair and using Eq.~\eqref{eq:appx_rope_fusion_commutes} gives
\begin{equation}
    R_h(p)^\top\widehat k_h(p)=\bar k_h(p).
    \label{eq:appx_rope_fusion_inverse}
\end{equation}
Thus, de-rotating the fused key is equivalent to de-rotating each source key before fusion.
In complex notation, the same operation multiplies each pair by \(e^{-\mathrm{i}\theta_r(p)}\), the conjugate of its original rotary factor.

\paragraph{Reapplying 1D RoPE.}
Let \(R_h^a\) denote the action module's 1D rotary transform, with its temporal axis, frequency grid, and channel pairing.
Expressing the video token's time \(t_p\) relative to the same temporal origin and in the same units as the action indices, we obtain
\begin{equation}
\begin{aligned}
    k_h^{v\to a}(p)
    &=R_h^a(t_p)R_h(p)^\top\widehat k_h(p)\\
    &=R_h^a(t_p)\bar k_h(p).
\end{aligned}
\label{eq:appx_rope_reapplication}
\end{equation}
Faster-WAM conditions only on the prefilled observed frame, for which \(t_p=0\).
Since \(R_h^a(0)=I\), reapplication is the identity and \(k_h^{v\to a}(p)=\bar k_h(p)\).
This operation removes the explicit video rotary factor from the keys supplied to the action module; their content can still encode spatial information.
The video backbone retains its native 3D RoPE.

Note that the retracted 1D-RoPE is not limited to WAMs condition only on the prefilled observed frame.
It can be directly applied to those condition on future video frames by assigning proper 1D-RoPE.

\subsection{Alternative RoPE Alignment: Lifted 3D RoPE}
\label{app:lifted_rope}
\label{sec:rope}

Besides retracted 1D RoPE, we also attempted an alternative solution by uplift the 1D RoPE of action tokens.
However, it introduces a stronger inductive bias on how action tokens attend to video tokens, 
which might hurt the generalizability according to our ablation study (Sec.~\ref{sec:ablations}).

Lifted 3D RoPE maps action timestamps into the video backbone's spatiotemporal coordinate system through spatial pseudo-anchors.

\paragraph{Spatial pseudo-anchors.}
We use the video backbone's temporal--height--width channel partition and frequency banks for action queries and keys.
Action token \(i\) and video token \(s\) retain their temporal indices \(i\) and \(t_s\), respectively.
Since action tokens have no image-grid coordinates, we assign them spatial pseudo-anchors at the camera-view centers.
Video keys retain their image-grid positions, so spatial rotary pairs encode offsets from these anchors.
For a head dimension of \(d=128\), the temporal, height, and width blocks contain \(22\), \(21\), and \(21\) rotary pairs, respectively, with
\(\omega^t_r=10000^{-2r/44}\) and
\(\omega^y_r=\omega^x_r=10000^{-2r/42}\), indexed from zero.
For a composite token grid of size \(G_y\times G_x\), the normalized center \((\bar c^y_c,\bar c^x_c)\) of camera view \(c\) gives
\begin{equation}
    c^y_c=G_y\bar c^y_c-\tfrac12,
    \qquad
    c^x_c=G_x\bar c^x_c-\tfrac12.
    \label{eq:dot_view_centers}
\end{equation}
\textcolor{black}{For action token \(i\), head \(h\), and video token \(s\) at \((t_s,y_s,x_s)\), the phases are}
\begin{equation}
\begin{aligned}
    \phi^{a,t}_{ihr}&=i\omega^t_r,
    &\phi^{v,t}_{sr}&=\textcolor{black}{t_s\omega^t_r},\\
    \phi^{a,y}_{ihr}&=c^y_{m_{hr}}\omega^y_r,
    &\phi^{v,y}_{sr}&=y_s\omega^y_r,\\
    \phi^{a,x}_{ihr}&=c^x_{m_{hr}}\omega^x_r,
    &\phi^{v,x}_{sr}&=x_s\omega^x_r,
\end{aligned}
\label{eq:dot_aligned_phases}
\end{equation}

Here, \(r\) ranges over the pairs of each axis, and \(m_{hr}\) selects a camera view as defined in Eq.~\eqref{eq:dot_two_view_schedule}.
For each head \(h\) and spatial frequency rank \(r\), the pseudo-anchor defines the coordinate lift \(i\mapsto(i,c^y_{m_{hr}},c^x_{m_{hr}})\).
The relative spatial phases are now
\(\omega^y_r(y_s-c^y_{m_{hr}})\) and
\(\omega^x_r(x_s-c^x_{m_{hr}})\), independent of the action index.
\textcolor{black}{The spatial rotary factors cancel in action--action attention because the pseudo-anchors are shared across action tokens.}
These assignments apply to attention in the action module; the video backbone retains its original 3D RoPE.
This design adds no learned positional parameters or additional attention operations.

\paragraph{Interleaving pseudo-anchors across views.}
Our Lifted 3D RoPE experiments on LIBERO and LIBERO-Plus use two camera views.
We interleave their pseudo-anchors across spatial frequency pairs, assigning pair \(r\) in head \(h\) to view
\begin{equation}
    m_{hr}=(r+h)\bmod 2,
    \label{eq:dot_two_view_schedule}
\end{equation}
where indices start at zero.
The two assignments form \(S_2\), comprising the identity and swap.
With \(H=24\), each view is assigned to \(12\) heads at every frequency index.
Height and width pairs share the same view assignment, while native frequencies and channel order are preserved.

\textcolor{black}{For multi-view settings such as RoboTwin, this construction could be extended using cyclic shifts or permutations of the view indices.
For \(C\) views, cyclic assignments take the form}
\begin{equation}
    m_{hr}=(r+h)\bmod C.
    \label{eq:dot_anchor_schedule}
\end{equation}
\textcolor{black}{Alternatively, one could use \(m_{hr}=\pi_h(r\bmod C)\), with \(\pi_h\) drawn from the permutation group \(S_C\).
Uniform replication requires \(C\mid H\) for cyclic shifts or \(C!\mid H\) for all permutations.
For RoboTwin's three views, these choices yield three cyclic shifts or six permutations in \(S_3\); either would assign each view to \(8\) of the \(24\) heads at every frequency index under uniform replication.}

\clearpage
\newpage
\section{Training and Inference}
\label{app:training_details}

\paragraph{Training objective.}
We use joint video and action flow-matching losses following Fast-WAM~\cite{yuan2026fastwam}.
Let \(X^v\) denote future video latents and \(X^a\) the demonstrated action chunk.
For each modality \(m\in\{v,a\}\), we independently sample standard Gaussian noise \(\epsilon^m\sim\mathcal N(0,I)\) and a noise level \(\sigma_m\in[0,1]\) from the schedule below, constructing
\begin{equation}
    X^m_{\sigma_m}
    =(1-\sigma_m)X^m+\sigma_m\epsilon^m,
    \qquad
    u^m=\epsilon^m-X^m.
    \label{eq:dot_flow_target}
\end{equation}
The corresponding DiT predicts the target velocity \(u^m\).
We minimize
\begin{equation}
    \mathcal L
    =\sum_{m\in\{v,a\}}\lambda_m\,
    \mathbb E\!\left[
        w_m(\sigma_m)
        \bigl\|f^m_\theta(X^m_{\sigma_m},\sigma_m;c_m)-u^m\bigr\|_{\mathrm{valid}}^2
    \right],
    \label{eq:dot_training_objective}
\end{equation}
where \(c_v\) contains the clean observed frame, language instruction, and proprioception, while \(c_a\) is the fused video KV context of the observed frame.
Here, \(\|\cdot\|_{\mathrm{valid}}^2\) denotes the mean squared error over non-padding target elements within each sample.
The observed frame remains clean and is excluded from the video prediction loss.
We use \(\lambda_v=\lambda_a=1\); the attention masks are specified in Sec.~\ref{sec:dot_training_inference}.

\paragraph{Noise schedule and weighting.}
For LIBERO and RoboTwin, we independently sample \(z_m\sim\mathcal U(0,1)\) for each modality and set \(\sigma_m=\phi(z_m)\), where \(\phi(z)=5z/(1+4z)\).
Both modalities use the same noise-level weighting function:
\[
    g(\sigma)=\exp\!\left[-2(\sigma-\tfrac12)^2\right]-\exp(-\tfrac12),
    \qquad
    w_v(\sigma)=w_a(\sigma)=\frac{g(\sigma)}{Z+10^{-10}},
\]
where \(Z=T^{-1}\sum_{r=1}^{T}g(\phi(r/T))\) and \(T=1000\).
Each DiT receives the rescaled timestep \(T\sigma_m\).

\paragraph{Optimization.}
We initialize the video backbone from Wan2.2-TI2V-5B and the action expert randomly.
We jointly optimize the video backbone, action expert, Lite KV-Fusion, and proprioceptive projection on downstream demonstrations.
The video autoencoder and text encoder remain fixed, and text embeddings are precomputed for training.
Gradients from the action loss propagate through Lite KV-Fusion into the video backbone.
Optimizer settings and benchmark-specific training configurations are provided in Appx.~\ref{app:experimental_optimization} and~\ref{app:full_reproducibility}.

\paragraph{Inference.}
For each action chunk, the video backbone performs a single prefill pass on the current observation, conditioned on the language instruction and proprioceptive state, without video denoising.
Lite KV-Fusion aggregates the native KV pairs across video layers, providing a fixed conditioning context throughout action denoising.
We apply retracted 1D-RoPE to the fused video keys before joint attention, as defined in Eq.~\eqref{eq:dot_retracted_rope}.
For LIBERO and RoboTwin, we predict \(32\)-action chunks using \(N=10\) explicit Euler steps with
\(\sigma_k=\phi(1-k/N)\), \(k=0,\ldots,N\).
Starting from \(X^a_{\sigma_0}\sim\mathcal N(0,I)\), we update
\[
    X^a_{\sigma_{k+1}}
    =
    X^a_{\sigma_k}
    +(\sigma_{k+1}-\sigma_k)
      f^a_\theta(X^a_{\sigma_k},\sigma_k;c_a).
\]
The final step reaches \(\sigma_N=0\).
Repeated denoising is confined to the action expert; no future video frames are generated for action prediction.
Benchmark-specific action-execution protocols are provided in Appx.~\ref{app:benchmark_protocols}.

\clearpage
\section{Experimental Details}
\label{app:experimental_details}

\subsection{Implementation Settings}
\label{app:implementation_settings}

\subsubsection{Model Configuration}
\label{app:experimental_model_configuration}
Faster-WAM uses the pretrained Wan2.2-TI2V-5B video backbone and a randomly initialized, single-layer action expert.
The video backbone has \(30\) layers, hidden width \(3072\), and FFN width \(14336\).
The action expert has hidden width \(1024\) and FFN width \(4096\).
Its query, key, and value projections map to an attention width of \(3072\), matching the video backbone's \(24\) heads of dimension \(128\); the attention output is projected back to width \(1024\).
The action expert uses retracted 1D-RoPE (Sec.~\ref{sec:action_head_design}) and has no text cross-attention. We use Lite KV-Fusion as described in Sec.~\ref{sec:kv_fusion}.

\subsubsection{Trainable and Frozen Modules}
\label{app:experimental_trainable_modules}
We jointly train the Wan2.2 video backbone (\(5.00\)B parameters), action expert (\(28.6\)M parameters), and Lite KV-Fusion (\(720\) parameters).
Lite KV-Fusion learns one layer-mixing logit for each attention head, video layer, and action layer, giving \(24\times30\times1=720\) parameters.
The same mixing weights are used for keys and values, so they are counted only once.
The proprioceptive projection is also trainable (\(36864\) parameters for LIBERO and \(61440\) for RoboTwin).
The VAE (\(704.69\)M parameters) remains frozen, and text embeddings are precomputed with the same text encoder as Fast-WAM~\cite{yuan2026fastwam}.
Appx.~\ref{app:training_details} describes the training objective and gradient flow.

\subsubsection{Optimization}
\label{app:experimental_optimization}
We use AdamW to optimize all trainable modules, including the proprioceptive projection, with a peak learning rate \(10^{-4}\), \(\beta=(0.9,0.95)\), weight decay \(0.01\), and gradient clipping at \(1.0\).
We use linear warmup for the first \(5\%\) of optimizer steps, followed by cosine decay to \(10^{-6}\).
Video and action noise levels are sampled independently as \(\sigma=5u/(1+4u)\), where \(u\sim\mathcal U(0,1)\).

\subsection{Benchmark Protocols}
\label{app:benchmark_protocols}

\subsubsection{LIBERO Training and Preprocessing}
\label{app:experimental_libero_training}
We use the preprocessed LIBERO~\cite{liu2023libero} dataset released by Fast-WAM~\cite{yuan2026fastwam}, generated with MuJoCo \(3.3.2\).
The dataset contains \(1712\) demonstrations with no-op actions removed, spanning \(40\) tasks and \(277713\) frames across the Spatial, Object, Goal, and Long suites.
We train jointly on all four suites.
Agent-view and wrist images are each resized to \(224\times224\) and concatenated horizontally to form a \(224\times448\) image.
Actions are \(7\)-dimensional and proprioception is \(8\)-dimensional.

We train for \(10\) epochs with a global batch size of \(128\).
Training comprises \(21700\) optimizer steps, including \(1085\) warmup steps.
We evaluate the final training checkpoint.

\subsubsection{LIBERO and LIBERO-Plus Evaluation}
\label{app:experimental_libero_evaluation}
Following the official LIBERO benchmark protocol, we evaluate \(50\) trials per task across \(40\) tasks in the four suites.
The policy predicts \(32\) actions and executes \(10\) before replanning.
We evaluate the same LIBERO-trained checkpoint on LIBERO-Plus~\cite{fei2026liberoplus} to assess generalizability under distribution shifts, using \(10030\) tasks across seven perturbation categories.

We evaluate with three seeds and report the mean and standard deviation of the success rate.

\subsubsection{RoboTwin Training and Preprocessing}
\label{app:experimental_robotwin_training}
Following the multi-task setup used by Fast-WAM~\cite{yuan2026fastwam}, Motus~\cite{bi2026motus}, and LingBot-VA~\cite{li2026causal}, we train jointly on clean and randomized RoboTwin~2.0 data~\cite{chen2025robotwin}.
The dataset comprises \(2500\) demonstrations collected in clean scenes and \(25000\) collected in randomized scenes.
The visual input consists of images from one head camera and two wrist cameras, combined into a single \(384\times320\) image.
Actions and proprioception are \(14\)-dimensional.

We train for \(5\) epochs with a global batch size of \(512\).
Training comprises \(58710\) optimizer steps, including \(2935\) warmup steps.
We evaluate the final training checkpoint.

\subsubsection{RoboTwin Evaluation and Action Execution}
\label{app:experimental_robotwin_evaluation}
Following prior work~\cite{yuan2026fastwam,bi2026motus,li2026causal}, we evaluate on \(50\) RoboTwin tasks under both clean and randomized conditions, using \(100\) trials per task.
We evaluate with three seeds and report the mean and standard deviation of the success rate.
The policy predicts \(32\) actions and executes \(24\) before replanning.
Actions are executed as joint-position and gripper commands, and success is determined by the environment's task-completion signal.
We report average success across clean and randomized conditions.

\subsection{Transfer to the LTX Video Backbone}
\label{app:ltx_backbone}

To test whether DoT transfers to other world action models beyond Wan2.2, we implement
Fast-LTX and Faster-LTX with LTX-Video-v0.9~\cite{hacohen2024ltx}, analogous to Fast-WAM and Faster-WAM.
We train two LTX variants on LIBERO and evaluate on both LIBERO and LIBERO-Plus.

\begin{table}[H]
    \centering
    \caption{\textbf{Transfer across video world models.}
    LIBERO and LIBERO-Plus success rates (\%) with the sizes of the video world models and depths of the action experts.
    Video size counts the video DiT parameters.
    LTX results use one evaluation run with both environment and action-noise seeds set to \(42\); Faster-WAM reports mean \(\!\pm\!\) std over three evaluation runs.
    Bold indicates the stronger result within each backbone family.}
    \label{tab:ltx_backbone_transfer}
    \setlength{\tabcolsep}{4pt}
    \resizebox{0.9\textwidth}{!}{
    \begin{tabular}{llcccc}
        \toprule
        \textbf{Method} &
        \shortstack{\textbf{Video world}\\\textbf{model}} &
        \shortstack{\textbf{Video}\\\textbf{params.}} &
        \shortstack{\textbf{Action}\\\textbf{depth}} &
        \shortstack{\textbf{LIBERO}\\\textbf{Avg.}} &
        \shortstack{\textbf{LIBERO-Plus}\\\textbf{Overall}} \\
        \midrule
        Fast-WAM & Wan2.2-TI2V-5B & \(5\)B & 30 & 97.6 & 51.5 \\
        \textbf{Faster-WAM} & Wan2.2-TI2V-5B & \(5\)B & 1 &
        \(\mathbf{98.3}\) & \(\mathbf{78.3}\) \\
        \midrule
        Fast-LTX & LTX-Video-v0.9 & \(1.92\)B & 28 & 95.1 & 43.0 \\
        \textbf{Faster-LTX} & LTX-Video-v0.9 & \(1.92\)B & 1 &
        \textbf{96.4} & \textbf{53.6} \\
        \bottomrule
    \end{tabular}}
\end{table}

Across both Wan2.2 and LTX-Video, DoT reduces the action expert to a single layer while improving LIBERO and LIBERO-Plus over the corresponding full-depth baseline, supporting the world-model-centric design across video backbones.

\clearpage
\section{Latency Reproducibility}
\label{app:latency_reproducibility}

\subsection{Scope and Latency Sources}
\label{app:latency_published_references}
We compare the model computation required to predict one action chunk, including text and vision encoding, multimodal conditioning, and action generation. We measure $\pi_0$, $\pi_{0.5}$, X-VLA, Fast-WAM, Faster-WAM, LingBot-VA, and Motus on the same 24\,GiB GPU model used to report GE-ACT's latency~\citep[Section~3.3]{liao2026genie}. The tables report the sum of the profiled model phases, excluding external input preparation, transfers, and robot execution. Fast-WAM, LingBot-VA, and Motus exceed the GPU's memory capacity as complete pipelines, so their totals combine separately timed components. The following subsections specify the implementations, configurations, and aggregation procedure.

LIBERO and RoboTwin use separate configurations, with the action-chunk length reported alongside each latency. Among the entries in our benchmark tables, GE-ACT is the sole published latency: we retain the $200$\,ms reported by Genie Envisioner for a 54-action prediction with one video denoising step and five action denoising steps~\citep[Section~3.3]{liao2026genie}, following the authors' measurement protocol.

\subsection{Implementations and Measurement Procedure}
\label{app:latency_measurement_procedure}

\subsubsection{Code and Upstream Checkouts}
\label{app:latency_implementations}
We use the following official implementations:
\begin{itemize}
    \item $\pi_0$ and $\pi_{0.5}$: OpenPI, \url{https://github.com/Physical-Intelligence/openpi}, upstream checkout \texttt{15a9616a0094}.
    \item X-VLA: \url{https://github.com/2toinf/X-VLA}, based on upstream checkout \texttt{6bc2513f5f1c}.
    \item Fast-WAM: \url{https://github.com/yuantianyuan01/FastWAM}, upstream checkout \texttt{45d8e1458921}.
    \item Motus: \url{https://github.com/thu-ml/Motus}, upstream checkout \texttt{f771216802b8a}.
    \item LingBot-VA: \url{https://github.com/Robbyant/lingbot-va}, upstream checkout \texttt{7c6ffa9bfc4b8}.
\end{itemize}
These checkouts identify the upstream sources; we add local benchmark wrappers and timing instrumentation.

\subsubsection{Hardware and Execution}
\label{app:latency_hardware}
We measure latency on a consumer GPU with 24\,GiB of memory.
We run one policy process at a time on a single GPU, with batch size one. 
\begin{itemize}
    \item \textbf{$\pi_0$ and $\pi_{0.5}$ (OpenPI).} JAX and jaxlib~0.5.3 with compiled inference and bfloat16 compute.
    \item \textbf{WAM (Fast-WAM and Faster-WAM).} PyTorch~2.7.1 and CUDA~12.8, using bfloat16 without \texttt{torch.compile}.
    \item \textbf{X-VLA.} PyTorch~2.1.2 and CUDA~12.1, using float32 without \texttt{torch.compile}.
    \item \textbf{LingBot-VA.} PyTorch~2.9.0 and CUDA~12.6, using bfloat16 model parameters without \texttt{torch.compile}.
    \item \textbf{Motus.} PyTorch~2.7.1 and CUDA~12.8, without \texttt{torch.compile}; the transformer components use bfloat16 with some float32 action-expert parameters, and the VAE parameters use float32.
\end{itemize}

\subsubsection{Timing Protocol}
\label{app:latency_timing_protocol}
We use 4,000 measured calls per component after five warmup calls. We report the arithmetic mean and standard deviation in milliseconds, with no outlier removal. For jointly measured phases, we compute the sample standard deviation of their per-call sum; for separately measured components, we combine sample variances as described below. Standard deviations describe variation across calls within the recorded run, including occasional latency spikes; they are not confidence intervals across independent runs.

\Needspace{4\baselineskip}
\subsection{Inference Latency and Phase Breakdown}
\label{app:latency_robotwin_phases}
\label{app:latency_libero_phases}

We benchmark policies using their LIBERO and RoboTwin configurations. Each measurement predicts one action chunk; Tabs.~\ref{tab:latency_statistics} and~\ref{tab:robotwin_latency_statistics} report the phase timings and action-chunk lengths for the two benchmarks. The vision encoder runs for every prediction, and text encoding or embedding is included; tokenization is excluded from timing.

\noindent\textbf{LIBERO.} $\pi_0$, $\pi_{0.5}$, X-VLA, Fast-WAM, and Faster-WAM use ten action denoising steps. LingBot-VA uses 20 video denoising steps and 50 action denoising steps, followed by one final cache-building forward for each modality; it is measured after its observation-history buffer is full.

\noindent\textbf{RoboTwin.} $\pi_0$, $\pi_{0.5}$, X-VLA, Fast-WAM, and Faster-WAM use ten action denoising steps. LingBot-VA uses 25 video denoising steps and 50 action denoising steps, followed by one final cache-building forward for each modality; it is measured after its observation-history buffer is full. Motus uses ten joint video/action denoising steps and retains native video decoding. Its Qwen vision--language model runs eleven times per action chunk.

Tab.~\ref{tab:robotwin_phase_definitions} defines the four inference phases used in both timing tables. Motus is included only in the RoboTwin comparison.
OpenPI phases are profiled as separately compiled functions with output synchronization. X-VLA uses CUDA events around the named phases and synchronizes each profiled call; the WAM component measurements use synchronized wall-clock timing. Profiling boundaries and separate compilation can make these totals differ from native model-call times.

For Faster-WAM, the video backbone is prefilled once per observation and the fused KV cache is reused across the ten action denoising steps. Both WAM implementations prepare the action attention mask once per prediction and reuse it across the denoising steps. Instruction encoding is included in every reported prediction.
\begin{table}[H]
\centering
\small
\setlength{\tabcolsep}{6pt}
\renewcommand{\arraystretch}{1.1}
\begin{tabular}{@{}p{0.22\linewidth}p{\dimexpr0.78\linewidth-2\tabcolsep\relax}@{}}
\toprule
Phase & Operations by model \\
\midrule
\raggedright Text encoding / embedding &
\raggedright OpenPI and X-VLA: token embedding.
WAM: UMT5-XXL encoding.
LingBot-VA: positive/negative UMT5-XXL encoding.
Motus: UMT5-XXL encoding and Qwen token embedding. \tabularnewline
\addlinespace[3pt]
\raggedright Vision encoder &
\raggedright OpenPI: SigLIP.
X-VLA: Florence-2 vision encoder and projection.
WAM and LingBot-VA: VAE encoding.
Motus: VAE encoding and Qwen vision encoding. \tabularnewline
\addlinespace[3pt]
\raggedright Multimodal conditioning &
\raggedright OpenPI: PaliGemma image--text prefix.
X-VLA: Florence-2 encoding of text and first view.
WAM: robot-state encoding and video prefill.
LingBot-VA: history-cache updates and video generation.
Motus: Qwen multimodal encoding and T5 projection. \tabularnewline
\addlinespace[3pt]
\raggedright Action generation &
\raggedright OpenPI and WAM: action denoising.
X-VLA: action denoising, additional-view fusion, setup, postprocessing, and instrumentation.
LingBot-VA: action denoising, final cache update, and remaining model work.
Motus: joint video/action denoising, video decoding, and remaining model work. \tabularnewline
\bottomrule
\end{tabular}
\captionsetup{font=small,skip=4pt}
\caption{Operations in each phase. OpenPI denotes $\pi_0$ and $\pi_{0.5}$; WAM denotes Fast-WAM and Faster-WAM.}
\label{tab:robotwin_phase_definitions}

\end{table}

\begin{table}[!htbp]
\centering
\small
\caption{LIBERO latency by phase. Values are mean $\pm$ standard deviation in milliseconds.}
\label{tab:latency_statistics}
\label{tab:libero_latency_phases}
\setlength{\tabcolsep}{3pt}
\resizebox{\linewidth}{!}{%
\begin{tabular}{@{}lrrrrrr@{}}
\toprule
Phase & $\pi_0$ & $\pi_{0.5}$ & X-VLA & LingBot-VA & Fast-WAM & Faster-WAM \\
\midrule
Text encoding / embedding & $0.205 \pm 0.043$ & $0.207 \pm 0.045$ & $0.016 \pm 0.002$ & $105.52 \pm 0.12$ & $17.48 \pm 0.03$ & $17.48 \pm 0.03$ \\
Vision encoder & $12.02 \pm 0.09$ & $12.01 \pm 0.08$ & $14.78 \pm 0.06$ & $68.31 \pm 0.31$ & $8.05 \pm 0.04$ & $8.09 \pm 0.03$ \\
Multimodal conditioning & $31.76 \pm 0.09$ & $35.08 \pm 0.07$ & $2.66 \pm 0.03$ & $1562.62 \pm 2.57$ & $20.38 \pm 0.05$ & $20.63 \pm 0.06$ \\
Action generation & $20.98 \pm 0.09$ & $20.42 \pm 0.07$ & $76.67 \pm 0.28$ & $3020.43 \pm 5.12$ & $150.12 \pm 1.65$ & $5.84 \pm 0.10$ \\
\midrule
Profiled phase total & $64.96 \pm 0.17$ & $67.72 \pm 0.14$ & $94.13 \pm 0.31$ & $4756.87 \pm 7.03$ & $196.03 \pm 1.67$ & $52.04 \pm 0.18$ \\
Action-chunk length & 50 & 10 & 30 & 16 & 32 & 32 \\
\bottomrule
\end{tabular}%
}
\end{table}

\begin{table}[!htbp]
\centering
\small
\caption{RoboTwin latency by phase. Values are mean $\pm$ standard deviation in milliseconds.}
\label{tab:robotwin_latency_statistics}
\label{tab:robotwin_wam_stages}
\label{tab:robotwin_vla_stages}
\setlength{\tabcolsep}{3pt}
\resizebox{\linewidth}{!}{%
\begin{tabular}{@{}lrrrrrrr@{}}
\toprule
Phase & $\pi_0$ & $\pi_{0.5}$ & X-VLA & Motus & LingBot-VA & Fast-WAM & Faster-WAM \\
\midrule
Text encoding / embedding & $0.215 \pm 0.034$ & $0.216 \pm 0.033$ & $0.015 \pm 0.002$ & $54.16 \pm 0.06$ & $105.44 \pm 0.21$ & $17.52 \pm 0.04$ & $17.52 \pm 0.04$ \\
Vision encoder & $12.01 \pm 0.08$ & $11.95 \pm 0.07$ & $20.71 \pm 0.05$ & $100.69 \pm 0.68$ & $111.82 \pm 0.47$ & $8.92 \pm 0.04$ & $8.94 \pm 0.04$ \\
Multimodal conditioning & $31.75 \pm 0.11$ & $35.11 \pm 0.07$ & $2.56 \pm 0.02$ & $115.06 \pm 2.93$ & $3061.35 \pm 12.56$ & $20.79 \pm 0.05$ & $20.97 \pm 0.15$ \\
Action generation & $20.72 \pm 0.10$ & $23.95 \pm 0.08$ & $76.62 \pm 0.80$ & $1110.35 \pm 2.04$ & $4107.50 \pm 7.11$ & $149.55 \pm 1.34$ & $5.78 \pm 0.12$ \\
\midrule
Profiled phase total & $64.70 \pm 0.21$ & $71.23 \pm 0.14$ & $99.91 \pm 0.83$ & $1380.27 \pm 4.24$ & $7386.11 \pm 15.76$ & $196.77 \pm 1.36$ & $53.20 \pm 0.26$ \\
Action-chunk length & 50 & 50 & 30 & 16 & 32 & 32 & 32 \\
\bottomrule
\end{tabular}%
}
\end{table}

The Fast-WAM and LingBot-VA totals in both tables, and the Motus total for RoboTwin, combine separately measured components. For LingBot-VA, we sum the text-encoding, VAE-encoding, and video/action-core times. For Motus, we sum the UMT5-XXL encoding time and the profiled phases of its full prediction model. These component estimates exclude module loading and transfers between CPU and GPU. For phases measured together, we sum their times within each call and compute the mean and sample standard deviation of these per-call totals, preserving covariance between phases. For separately measured components, we sum their means and combine their standard deviations as $s_{\mathrm{total}}=\sqrt{\sum_i s_i^2}$, assuming zero covariance between components.

\begin{figure}[t]
    \centering
    \includegraphics[width=0.55\linewidth]{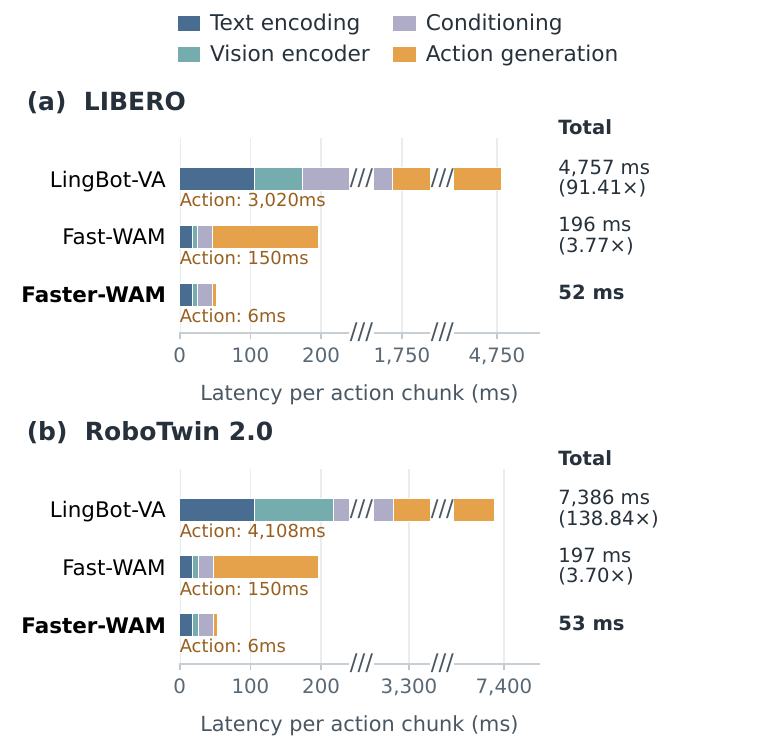}
    \caption{\textbf{Inference latency by phase.} ``Action'' labels show action-generation time; right-hand labels show total latency per action chunk, with ratios relative to Faster-WAM in parentheses.}
    \label{fig:inference_latency_phases}
\end{figure}

\Needspace{20\baselineskip}
\subsection{Inference Memory}
\label{app:latency_memory}

\begin{wraptable}{r}{0.44\textwidth}
\vspace{-6pt}
\centering
\small
\setlength{\tabcolsep}{6pt}
\begin{tabular}{@{}lr@{}}
\toprule
Policy & Peak memory (GiB) \\
\midrule
$\pi_0$ & 6.43\textsuperscript{$\dagger$} \\
$\pi_{0.5}$ & 6.65\textsuperscript{$\dagger$} \\
X-VLA & 3.46 \\
Fast-WAM & Does not fit \\
LingBot-VA & Does not fit \\
Motus & Does not fit \\
\midrule
Faster-WAM & 21.44 \\
\bottomrule
\end{tabular}
\captionsetup{font=small,skip=4pt}
\caption{Peak GPU tensor memory for RoboTwin with all policy components resident on the 24\,GiB device.}
\label{tab:robotwin_resident_memory}
\label{tab:robotwin_runtime_memory}
\par\smallskip
\begin{minipage}{\linewidth}
\footnotesize
\textsuperscript{$\dagger$}JAX process-wide peak, including loading, compilation, and warmup.
\end{minipage}
\end{wraptable}

We measure peak GPU tensor memory using the RoboTwin configurations, with all policy components resident, including the text encoder. The measurement includes model parameters, caches, and working tensors. For PyTorch, we discard five warmup calls and report the maximum over ten inference calls, synchronizing the GPU and resetting the peak counter before each call. For $\pi_0$ and $\pi_{0.5}$, JAX reports a process-wide peak that also includes loading, compilation, and warmup. Tab.~\ref{tab:robotwin_runtime_memory} summarizes the results. We report memory in GiB, where 1\,GiB equals \(2^{30}\) bytes.

Fast-WAM runs out of memory while loading its action expert with the text encoder already resident. LingBot-VA's model and required history cache together need at least $29.41$\,GiB, exceeding device capacity before working tensors are allocated. Motus's registered model storage alone requires $28.16$\,GiB. We therefore measure these models in separate components for the latency results above; no full-policy inference-memory peak is available for these three models.

\clearpage
\section{Real-Robot Experimental Setup}
\label{app:real_robot_setup}

\begin{figure}[htbp]
    \centering
    \includegraphics[width=0.88\linewidth]{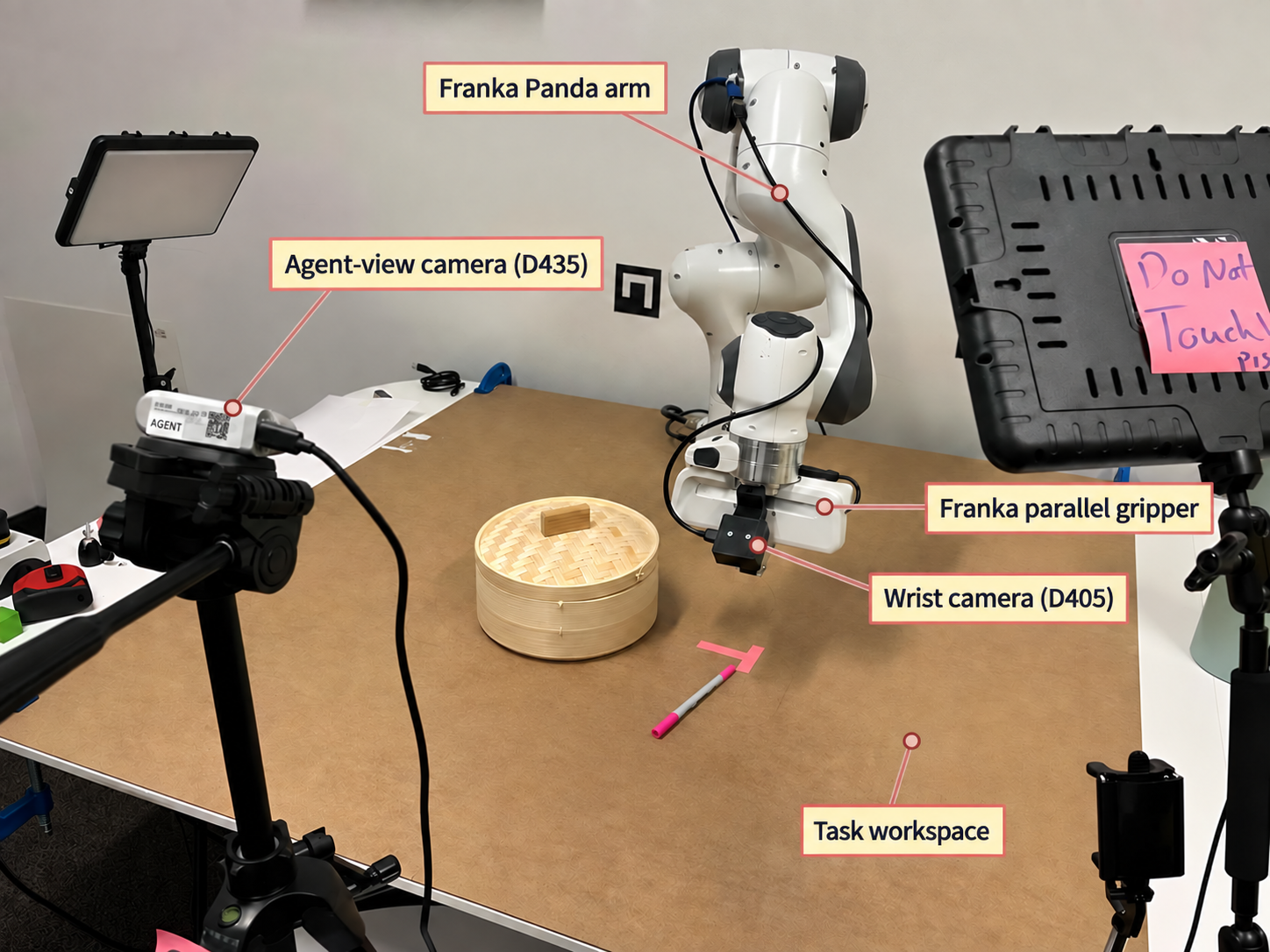}
    \caption{Real-robot experimental setup, shown with the fruit-picking task. Callouts identify the Franka Panda arm, parallel gripper, wrist-mounted D405, fixed agent-view D435, and tabletop workspace.}
    \label{fig:real_robot_setup}
\end{figure}

\subsection{Hardware and Control}
Fig.~\ref{fig:real_robot_setup} shows our Franka Panda arm with a Franka parallel gripper mounted at a tabletop workspace, facing a fixed agent-view camera.
Two Intel RealSense cameras provide complementary views of the scene: a wrist-mounted D405 observes the region near the gripper, while a fixed D435 captures the robot and task workspace.
The D405 provides an RGB field of view of \(87^\circ\times58^\circ\) (horizontal \(\times\) vertical).\footnote{Manufacturer specifications: \url{https://www.realsenseai.com/product-family/d405-series/}.}
The D435 provides an RGB field of view of \(69^\circ\times42^\circ\).\footnote{Manufacturer specifications: \url{https://www.realsenseai.com/products/stereo-depth-camera-d435/}.}
These specifications describe the camera views before image preprocessing.

The robot and cameras connect to a workstation running Linux with a real-time kernel.
Each policy receives RGB images from both cameras, robot proprioception, and a task-specific text prompt.
At each observation update, we read the currently available camera images and robot state.
The policy produces end-effector delta commands expressed in the tool frame, together with a gripper-state command.
Actions are predicted in chunks of 32, of which the first eight are executed before replanning from a new observation.
Polymetis\footnote{\url{https://facebookresearch.github.io/fairo/polymetis/}.} provides the control interface through which these commands are sent to the robot via the Franka Control Interface.

\subsection{Task Environments}
The four tasks use tabletop objects arranged to test different manipulation requirements.
In \texttt{place\_cube}, a red wooden cube and an IKEA bowl provide a basic pick-and-place setting.
The \texttt{pick\_fruit} task extends this setting to sequential manipulation: fruit objects are placed inside a lidded bamboo steamer, and the robot must lift the lid, place it aside, and pick the initially concealed plum.
For \texttt{push\_T}, two wooden blocks arranged in a T shape provide a contact-sensitive pushing task.
Finally, \texttt{dish\_utensil} uses standard IKEA dishes and utensils to test organization over a longer sequence of placements, with success determined by the final arrangement.

\subsection{Demonstration Collection}
We collect demonstrations with a SpaceMouse, which allows the operator to control the end effector in six degrees of freedom through the same Polymetis interface used for policy execution.
During teleoperation, camera observations, robot state, end-effector pose, and gripper state are recorded at 10\,Hz.
The resulting dataset contains 100 demonstrations per task, for a total of 400 demonstrations.

To introduce variation around each nominal task configuration, we change object placements between demonstrations.
For \texttt{place\_cube}, the cube's initial position varies within a \(10\,\mathrm{cm}\times10\,\mathrm{cm}\) region; for \texttt{pick\_fruit}, the plum is placed at different locations inside the container.
We apply smaller placement perturbations to \texttt{push\_T} and \texttt{dish\_utensil}, including variations in utensil arrangement, because these tasks already impose greater demands on contact control and sequential execution.

\subsection{Training and Evaluation}
Using the combined demonstration dataset, we train policies jointly across the four tasks with language conditioning and follow each model's recommended training recipe.
For Faster-WAM and Fast-WAM, this entails 30 training epochs.
We evaluate each policy over 20 rollouts per task and report the fraction of successful rollouts.

For \texttt{dish\_utensil}, we report a single task success rate based on its final checkpoint, which requires the complete dish-and-utensil arrangement to be correct.
The rate is the number of successful rollouts divided by 20, or by 10 for LingBot-VA.
Tab.~\ref{tab:real_world_results} summarizes the results, while Appx.~\ref{app:real_world_rollouts} provides timestamped examples of physical execution.

\clearpage
\section{Real-World Demo Rollout (Speed Comparison)}
\label{app:real_world_rollouts}

\subsection{Rollout Setup}
\label{app:real_world_rollout_setup}

We compare four real-robot rollouts of the \texttt{pick\_fruit} task, with the prompt
\emph{``Lift the lid, put it aside, and pick the black plum.''}
Columns identify the policies; rows sample elapsed wall-clock time from each
recording's start, including pauses between movements. We show 5-second intervals
through 40 seconds. Later snapshots use 20-second intervals while waiting for
LingBot-VA. Tab.~\ref{tab:real_world_rollout_times} reports the completion time of each rollout.

\begin{table}[H]
\centering
\small
\caption{Completion time of the \texttt{pick\_fruit} rollouts in Fig.~\ref{fig:real_world_speed_comparison}, measured as elapsed wall-clock time from the start of each recording.}
\label{tab:real_world_rollout_times}
\begin{tabular}{@{}lc@{}}
\toprule
Policy & Completion time (s) \\
\midrule
Faster-WAM & 21.2 \\
$\pi_{0.5}$ & 21.6 \\
Fast-WAM & 38.5 \\
LingBot-VA & 139.9 \\
\bottomrule
\end{tabular}
\end{table}

\subsection{Rollout Comparison}
\label{app:real_world_rollout_comparison}

\begingroup
\setlength{\tabcolsep}{2pt}
\definecolor{rolloutdone}{RGB}{0,112,76}
\definecolor{rolloutactive}{RGB}{85,90,96}
\newcommand{\rolloutframe}[2]{\begin{minipage}[c]{0.215\linewidth}\centering
  \setlength{\fboxsep}{0pt}\setlength{\fboxrule}{1.2pt}%
  \ifnum#2>0\relax
    \colorbox{rolloutdone}{\parbox[c][12pt][c]{\linewidth}{\centering
      \color{white}\fontsize{7}{8}\selectfont\bfseries TASK COMPLETE}}\par\nointerlineskip
    \fcolorbox{rolloutdone}{white}{#1}\par
    {\scriptsize\ifnum#2=1 Final frame held\else Recorded at 21 s\fi}%
  \else
    \colorbox{rolloutactive}{\parbox[c][12pt][c]{\linewidth}{\centering
      \color{white}\fontsize{7}{8}\selectfont IN PROGRESS}}\par\nointerlineskip
    \fcolorbox{rolloutactive}{white}{#1}\par
    {\scriptsize\strut}%
  \fi\end{minipage}}
\captionsetup{font=small,hypcap=false}
\captionof{figure}{\textbf{Real-world rollout speed comparison.}
Snapshots at 0, 5, 10, 15, 20, 25, 30, 35, and 40 seconds, then every 20 seconds to 140 seconds.
Faster-WAM and $\pi_{0.5}$ have both lifted the plum by the 25-second snapshot.
The sampling interval increases after Faster-WAM, $\pi_{0.5}$, and Fast-WAM have finished.
Green \textbf{TASK COMPLETE} banners and borders mark completed rollouts.
The final recorded frame is held in later snapshots; gray banners indicate ongoing execution.
These are illustrative rollouts, not estimates of mean task-completion time.}
\label{fig:real_world_speed_comparison}
\medskip
\begin{longtable}{@{}r*{4}{c}@{}}
\small Time & \small\textbf{Faster-WAM} & \small $\pi_{0.5}$ & \small Fast-WAM & \small LingBot-VA \\[4pt]
\endfirsthead
\small Time & \small\textbf{Faster-WAM} & \small $\pi_{0.5}$ & \small Fast-WAM & \small LingBot-VA \\[4pt]
\endhead
  0 s
  & \rolloutframe{\includegraphics[width=\dimexpr\linewidth-2\fboxrule\relax]{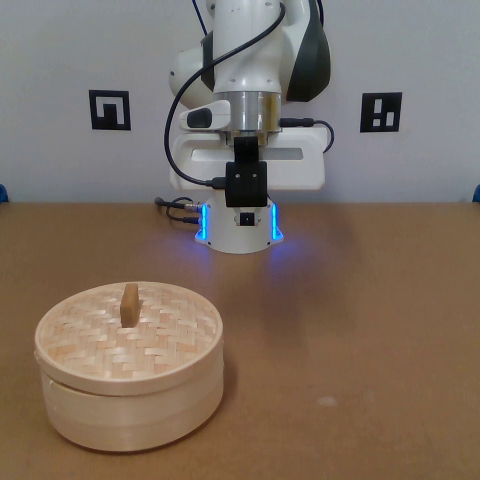}}{0}
  & \rolloutframe{\includegraphics[width=\dimexpr\linewidth-2\fboxrule\relax]{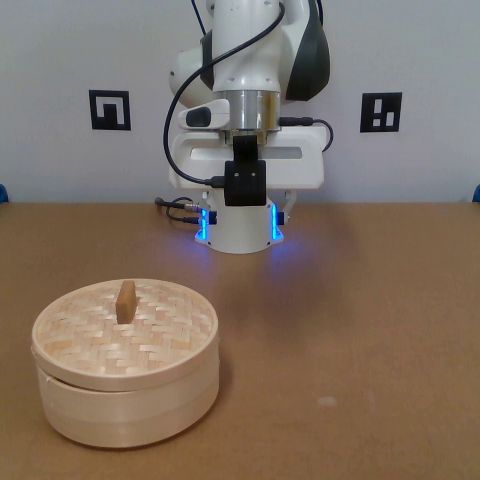}}{0}
  & \rolloutframe{\includegraphics[width=\dimexpr\linewidth-2\fboxrule\relax]{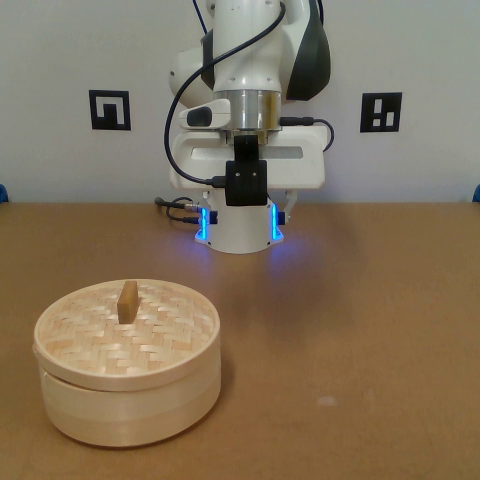}}{0}
  & \rolloutframe{\includegraphics[width=\dimexpr\linewidth-2\fboxrule\relax]{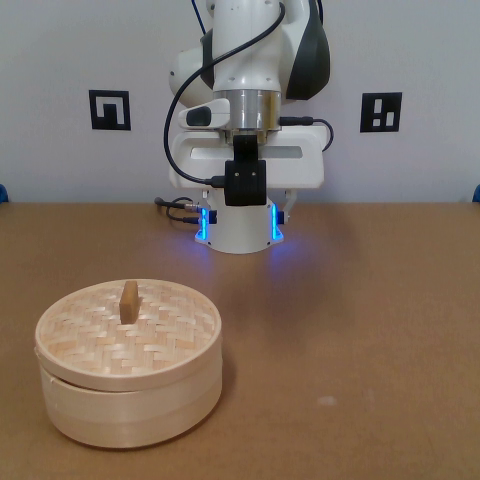}}{0}
  \\[4pt]
  5 s
  & \rolloutframe{\includegraphics[width=\dimexpr\linewidth-2\fboxrule\relax]{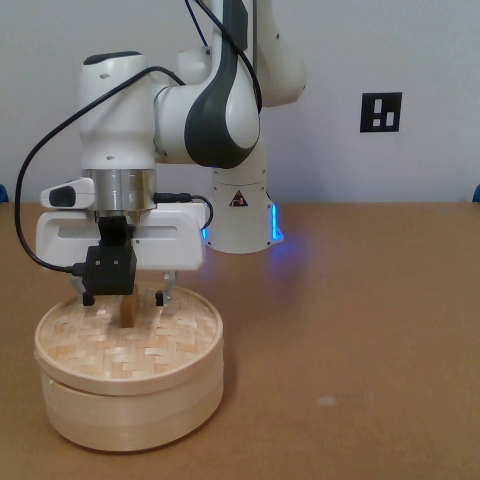}}{0}
  & \rolloutframe{\includegraphics[width=\dimexpr\linewidth-2\fboxrule\relax]{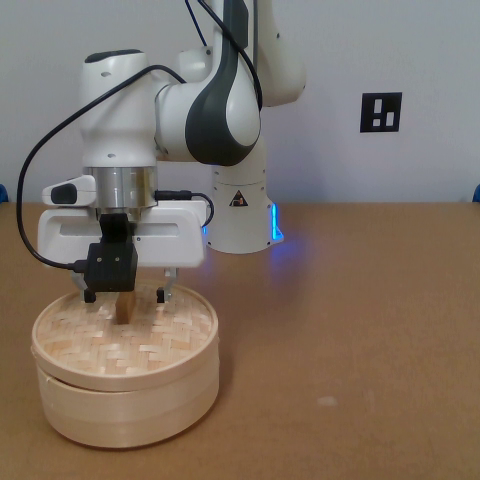}}{0}
  & \rolloutframe{\includegraphics[width=\dimexpr\linewidth-2\fboxrule\relax]{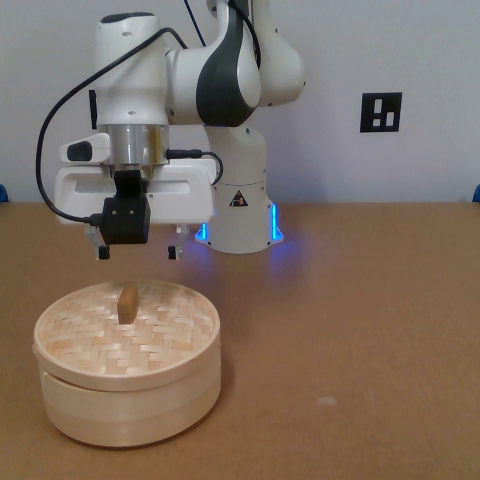}}{0}
  & \rolloutframe{\includegraphics[width=\dimexpr\linewidth-2\fboxrule\relax]{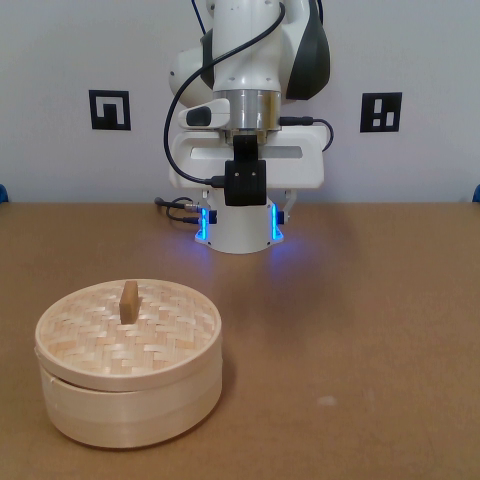}}{0}
  \\[4pt]
  10 s
  & \rolloutframe{\includegraphics[width=\dimexpr\linewidth-2\fboxrule\relax]{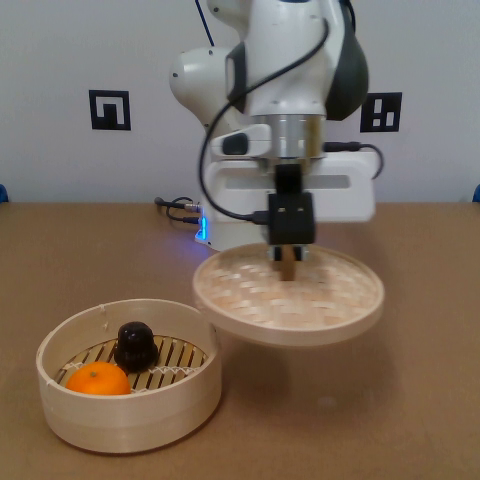}}{0}
  & \rolloutframe{\includegraphics[width=\dimexpr\linewidth-2\fboxrule\relax]{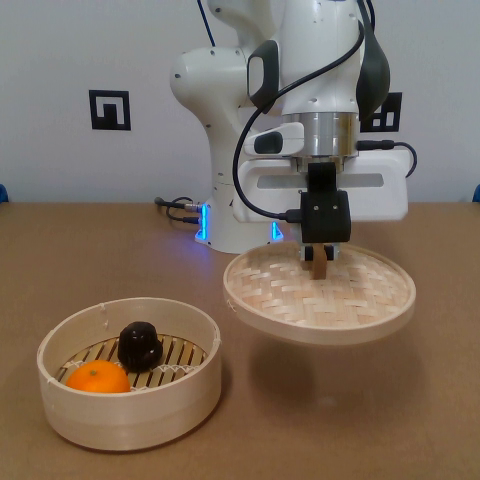}}{0}
  & \rolloutframe{\includegraphics[width=\dimexpr\linewidth-2\fboxrule\relax]{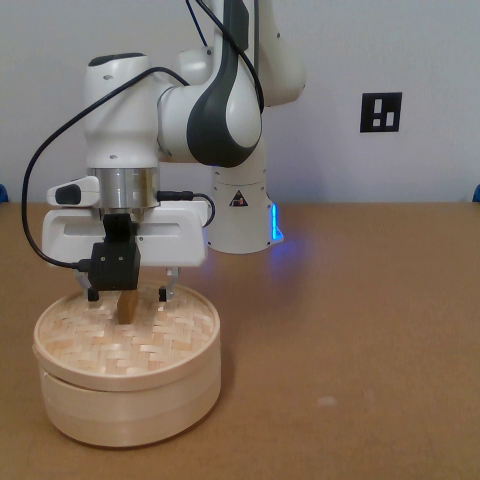}}{0}
  & \rolloutframe{\includegraphics[width=\dimexpr\linewidth-2\fboxrule\relax]{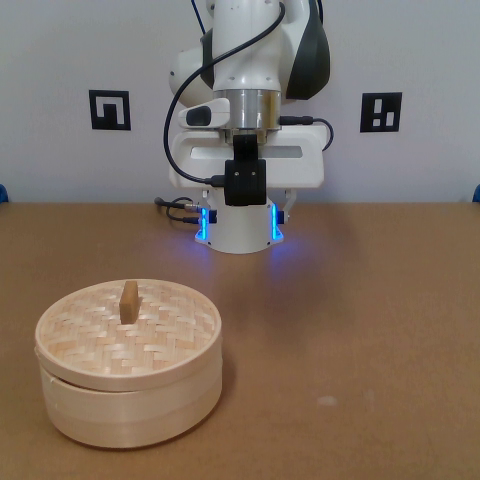}}{0}
  \\[4pt]
  15 s
  & \rolloutframe{\includegraphics[width=\dimexpr\linewidth-2\fboxrule\relax]{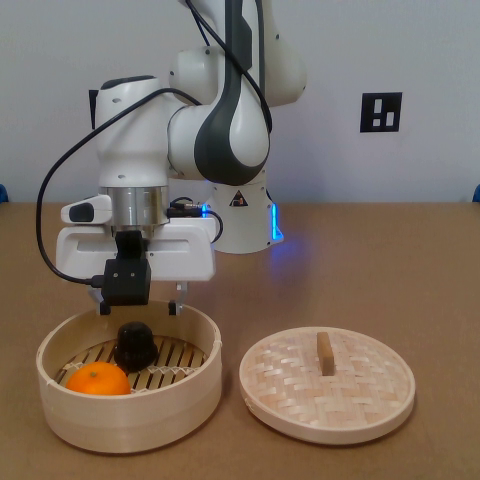}}{0}
  & \rolloutframe{\includegraphics[width=\dimexpr\linewidth-2\fboxrule\relax]{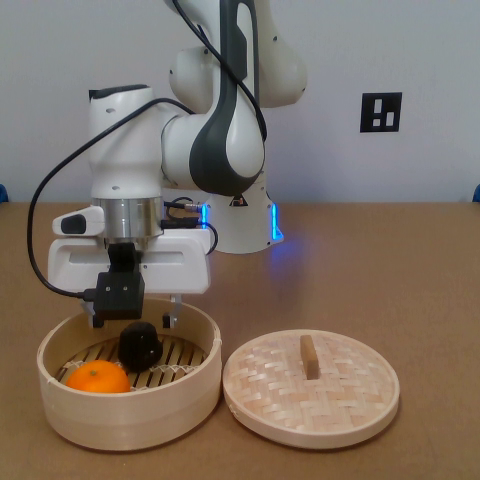}}{0}
  & \rolloutframe{\includegraphics[width=\dimexpr\linewidth-2\fboxrule\relax]{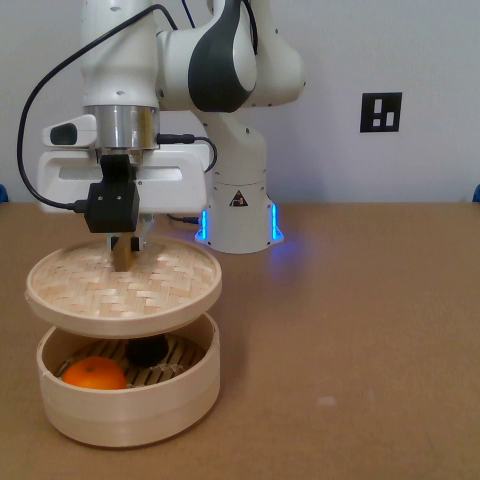}}{0}
  & \rolloutframe{\includegraphics[width=\dimexpr\linewidth-2\fboxrule\relax]{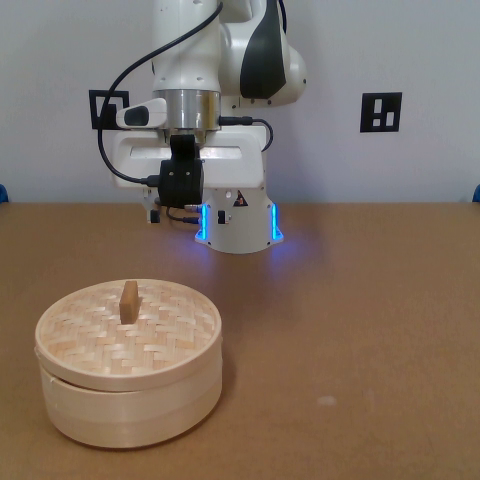}}{0}
  \\[4pt]
  20 s
  & \rolloutframe{\includegraphics[width=\dimexpr\linewidth-2\fboxrule\relax]{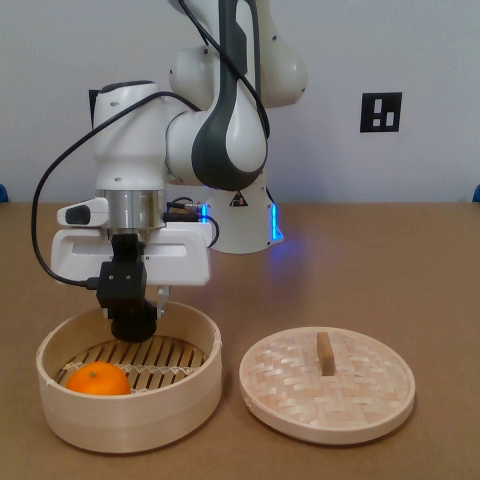}}{0}
  & \rolloutframe{\includegraphics[width=\dimexpr\linewidth-2\fboxrule\relax]{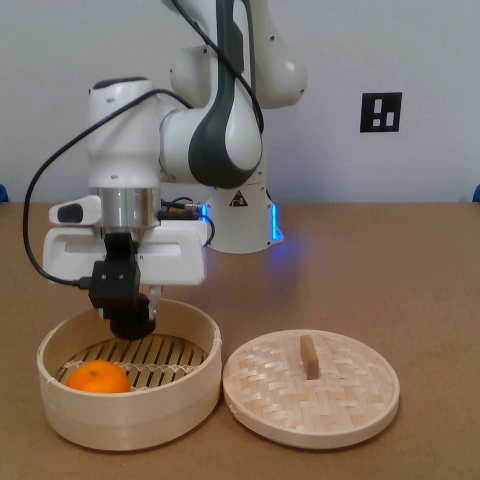}}{0}
  & \rolloutframe{\includegraphics[width=\dimexpr\linewidth-2\fboxrule\relax]{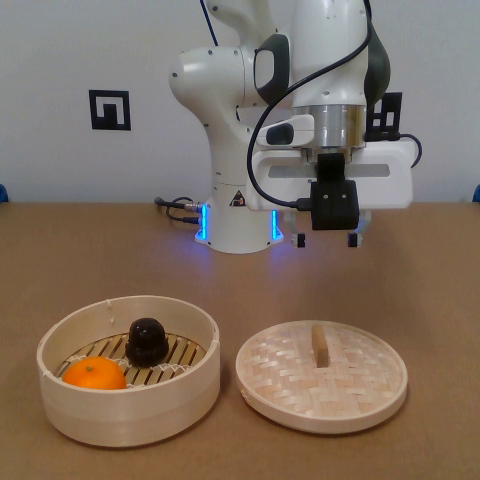}}{0}
  & \rolloutframe{\includegraphics[width=\dimexpr\linewidth-2\fboxrule\relax]{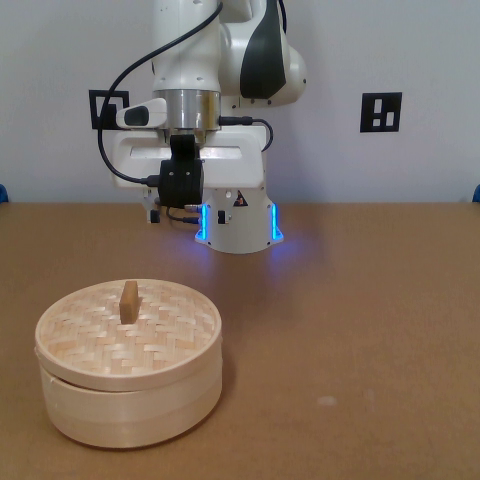}}{0}
  \\[4pt]
  25 s
  & \rolloutframe{\includegraphics[width=\dimexpr\linewidth-2\fboxrule\relax]{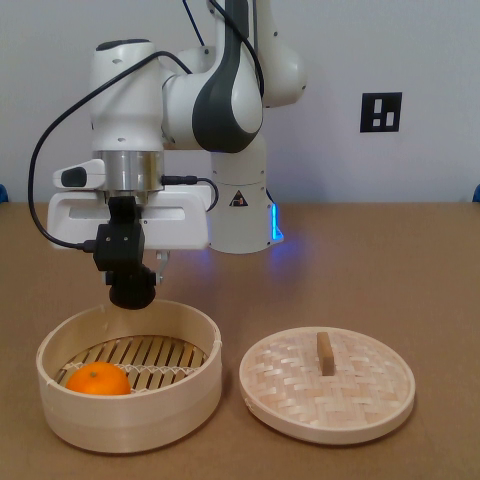}}{1}
  & \rolloutframe{\includegraphics[width=\dimexpr\linewidth-2\fboxrule\relax]{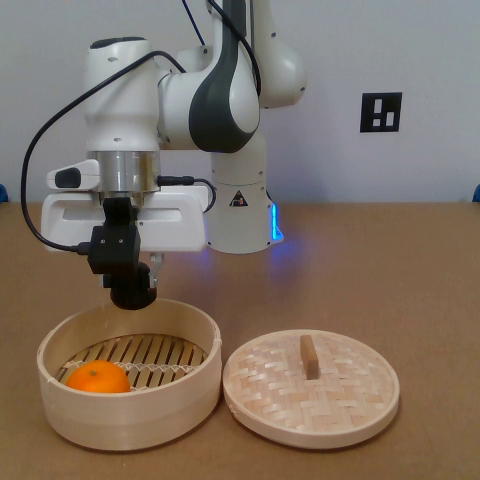}}{1}
  & \rolloutframe{\includegraphics[width=\dimexpr\linewidth-2\fboxrule\relax]{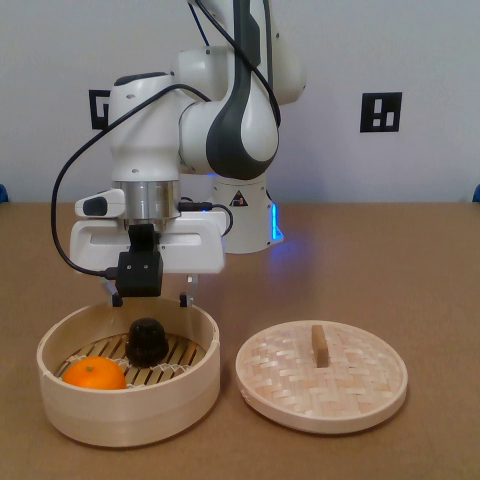}}{0}
  & \rolloutframe{\includegraphics[width=\dimexpr\linewidth-2\fboxrule\relax]{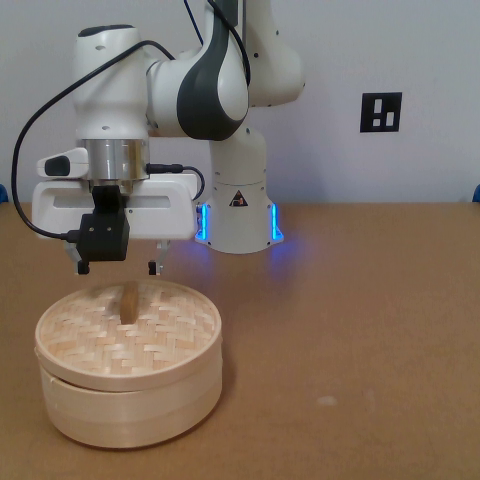}}{0}
  \\[4pt]
  30 s
  & \rolloutframe{\includegraphics[width=\dimexpr\linewidth-2\fboxrule\relax]{appx/real_world_rollouts/fasterwam/025s.png}}{1}
  & \rolloutframe{\includegraphics[width=\dimexpr\linewidth-2\fboxrule\relax]{appx/real_world_rollouts/pi05/025s.png}}{1}
  & \rolloutframe{\includegraphics[width=\dimexpr\linewidth-2\fboxrule\relax]{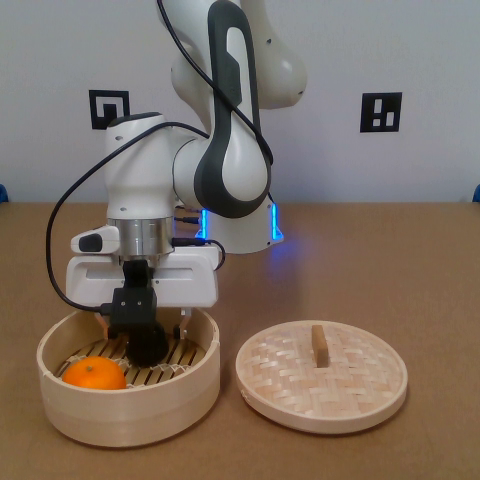}}{0}
  & \rolloutframe{\includegraphics[width=\dimexpr\linewidth-2\fboxrule\relax]{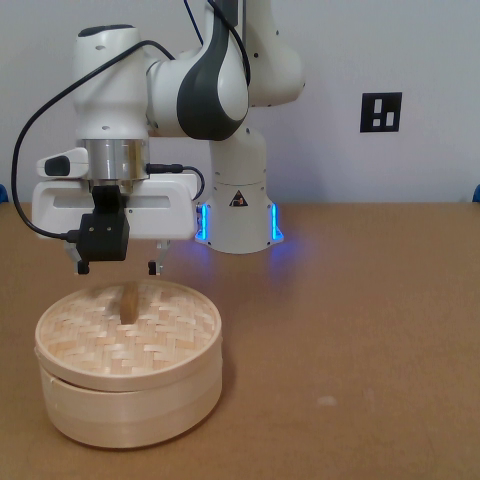}}{0}
  \\[4pt]
  35 s
  & \rolloutframe{\includegraphics[width=\dimexpr\linewidth-2\fboxrule\relax]{appx/real_world_rollouts/fasterwam/025s.png}}{1}
  & \rolloutframe{\includegraphics[width=\dimexpr\linewidth-2\fboxrule\relax]{appx/real_world_rollouts/pi05/025s.png}}{1}
  & \rolloutframe{\includegraphics[width=\dimexpr\linewidth-2\fboxrule\relax]{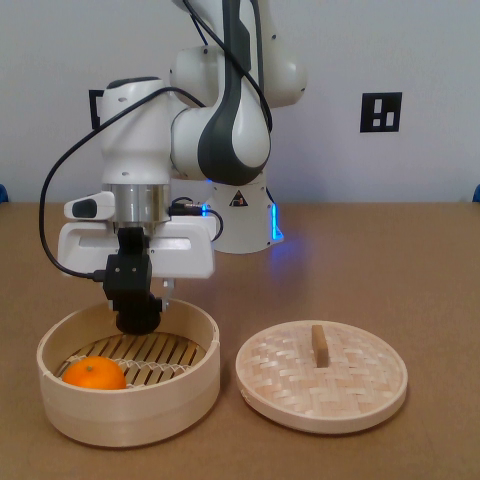}}{0}
  & \rolloutframe{\includegraphics[width=\dimexpr\linewidth-2\fboxrule\relax]{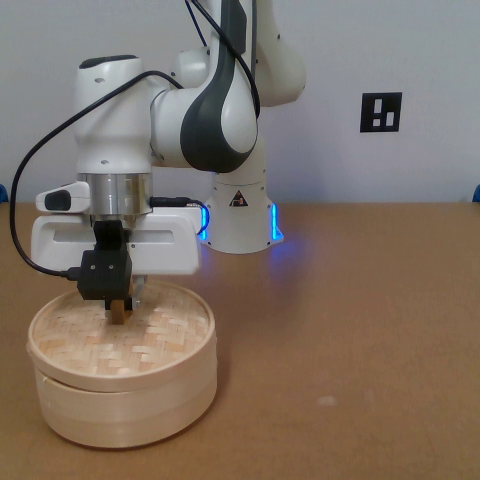}}{0}
  \\[4pt]
  40 s
  & \rolloutframe{\includegraphics[width=\dimexpr\linewidth-2\fboxrule\relax]{appx/real_world_rollouts/fasterwam/025s.png}}{1}
  & \rolloutframe{\includegraphics[width=\dimexpr\linewidth-2\fboxrule\relax]{appx/real_world_rollouts/pi05/025s.png}}{1}
  & \rolloutframe{\includegraphics[width=\dimexpr\linewidth-2\fboxrule\relax]{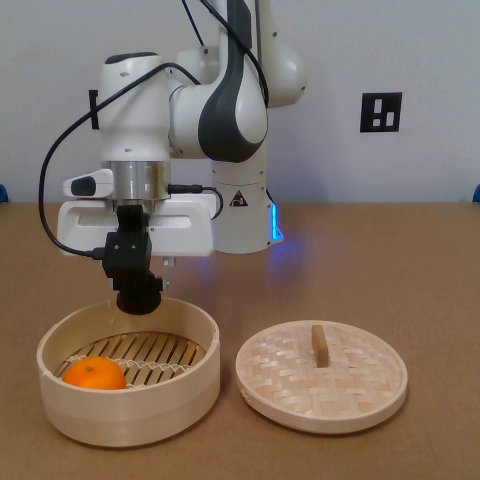}}{1}
  & \rolloutframe{\includegraphics[width=\dimexpr\linewidth-2\fboxrule\relax]{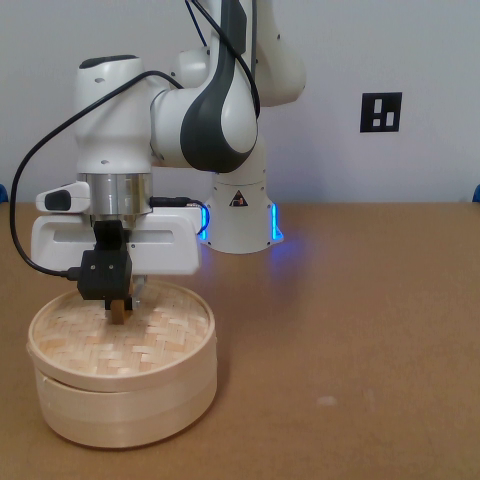}}{0}
  \\[4pt]
  $\vdots$ & $\vdots$ & $\vdots$ & $\vdots$ & $\vdots$ \\[4pt]
  60 s
  & \rolloutframe{\includegraphics[width=\dimexpr\linewidth-2\fboxrule\relax]{appx/real_world_rollouts/fasterwam/025s.png}}{1}
  & \rolloutframe{\includegraphics[width=\dimexpr\linewidth-2\fboxrule\relax]{appx/real_world_rollouts/pi05/025s.png}}{1}
  & \rolloutframe{\includegraphics[width=\dimexpr\linewidth-2\fboxrule\relax]{appx/real_world_rollouts/fastwam/040s.png}}{1}
  & \rolloutframe{\includegraphics[width=\dimexpr\linewidth-2\fboxrule\relax]{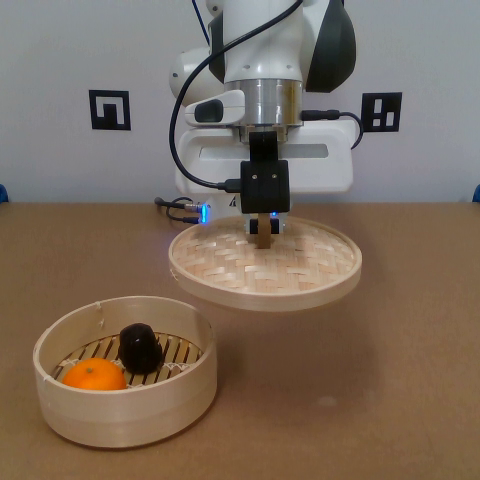}}{0}
  \\[4pt]
  80 s
  & \rolloutframe{\includegraphics[width=\dimexpr\linewidth-2\fboxrule\relax]{appx/real_world_rollouts/fasterwam/025s.png}}{1}
  & \rolloutframe{\includegraphics[width=\dimexpr\linewidth-2\fboxrule\relax]{appx/real_world_rollouts/pi05/025s.png}}{1}
  & \rolloutframe{\includegraphics[width=\dimexpr\linewidth-2\fboxrule\relax]{appx/real_world_rollouts/fastwam/040s.png}}{1}
  & \rolloutframe{\includegraphics[width=\dimexpr\linewidth-2\fboxrule\relax]{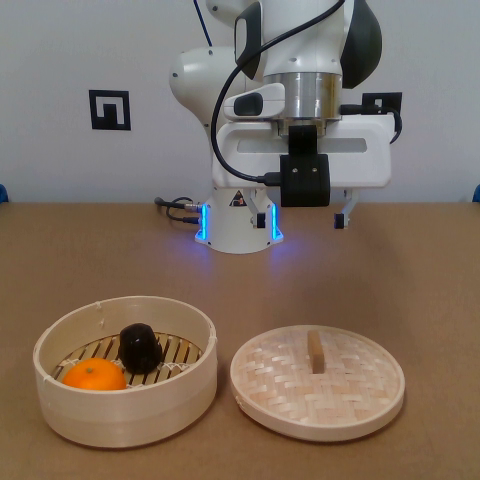}}{0}
  \\[4pt]
  100 s
  & \rolloutframe{\includegraphics[width=\dimexpr\linewidth-2\fboxrule\relax]{appx/real_world_rollouts/fasterwam/025s.png}}{1}
  & \rolloutframe{\includegraphics[width=\dimexpr\linewidth-2\fboxrule\relax]{appx/real_world_rollouts/pi05/025s.png}}{1}
  & \rolloutframe{\includegraphics[width=\dimexpr\linewidth-2\fboxrule\relax]{appx/real_world_rollouts/fastwam/040s.png}}{1}
  & \rolloutframe{\includegraphics[width=\dimexpr\linewidth-2\fboxrule\relax]{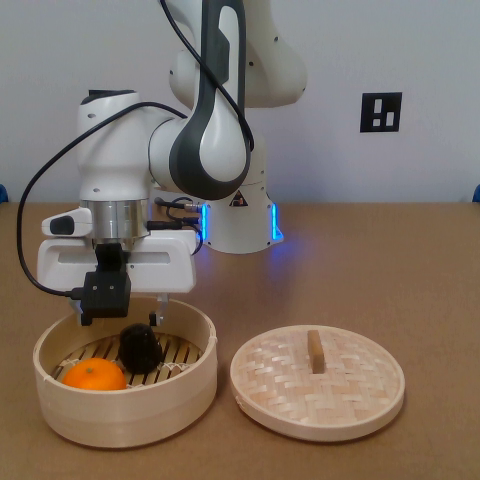}}{0}
  \\[4pt]
  120 s
  & \rolloutframe{\includegraphics[width=\dimexpr\linewidth-2\fboxrule\relax]{appx/real_world_rollouts/fasterwam/025s.png}}{1}
  & \rolloutframe{\includegraphics[width=\dimexpr\linewidth-2\fboxrule\relax]{appx/real_world_rollouts/pi05/025s.png}}{1}
  & \rolloutframe{\includegraphics[width=\dimexpr\linewidth-2\fboxrule\relax]{appx/real_world_rollouts/fastwam/040s.png}}{1}
  & \rolloutframe{\includegraphics[width=\dimexpr\linewidth-2\fboxrule\relax]{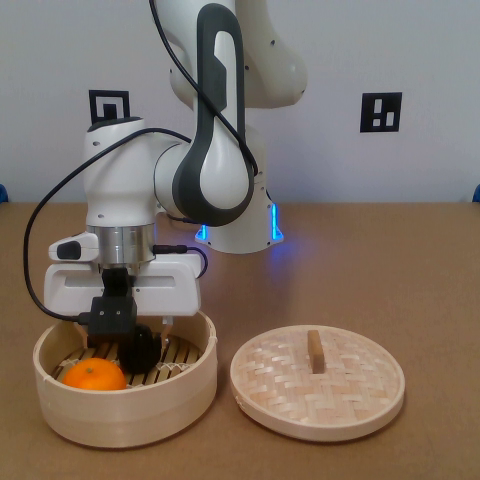}}{0}
  \\[4pt]
  140 s
  & \rolloutframe{\includegraphics[width=\dimexpr\linewidth-2\fboxrule\relax]{appx/real_world_rollouts/fasterwam/025s.png}}{1}
  & \rolloutframe{\includegraphics[width=\dimexpr\linewidth-2\fboxrule\relax]{appx/real_world_rollouts/pi05/025s.png}}{1}
  & \rolloutframe{\includegraphics[width=\dimexpr\linewidth-2\fboxrule\relax]{appx/real_world_rollouts/fastwam/040s.png}}{1}
  & \rolloutframe{\includegraphics[width=\dimexpr\linewidth-2\fboxrule\relax]{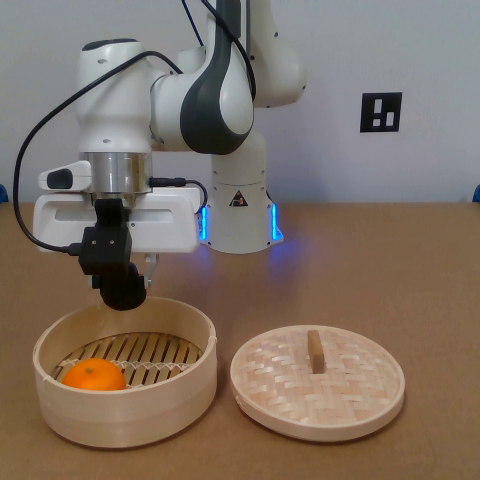}}{1}
  \\[4pt]
\end{longtable}
\endgroup

\clearpage
\clearpage
\section{Full Reproducibility Settings}
\label{app:full_reproducibility}
Tables~\ref{tab:fasterwam_libero_reproducibility} and~\ref{tab:fasterwam_robotwin_reproducibility}
summarize the Faster-WAM training recipe for LIBERO and RoboTwin.
Image sizes are given as height $\times$ width.
\subsection{LIBERO}
\label{app:reproducibility_libero}

\begin{table}[!htbp]
\centering
\footnotesize
\setlength{\tabcolsep}{4pt}
\renewcommand{\arraystretch}{1.05}
\caption{Faster-WAM training settings for LIBERO.}
\label{tab:fasterwam_libero_reproducibility}
\begin{tabularx}{\textwidth}{@{}l>{\raggedright\arraybackslash}X@{}}
\toprule Setting & Value \\
\midrule
\multicolumn{2}{@{}l}{\textbf{Training}} \\
\addlinespace[0.25em]
Epochs / optimization steps & 10 / 21,700 \\
Global batch size & 128 \\
Batch per GPU / gradient accumulation & 8 / 1 \\
Optimizer / betas & AdamW / $(0.9,\,0.95)$ \\
Peak / final learning rate & $10^{-4}$ / $10^{-6}$ \\
Schedule / linear warmup & Cosine / 5\% of training steps \\
Weight decay / gradient clipping & 0.01 / 1.0 \\
Precision / random seed & BF16 / 42 \\
Video backbone & Jointly fine-tuned \\
\addlinespace[0.8em]
\multicolumn{2}{@{}l}{\textbf{Data and preprocessing}} \\
\addlinespace[0.25em]
Training data & LIBERO Spatial, Object, Goal, Long; no-op actions removed \\
Camera views & External and wrist \\
Final image size per camera & External: $224\times224$; wrist: $224\times224$ \\
Model input image size & $224\times448$ \\
Sequence length / sampling stride & 33 / 1 \\
Action-to-video frequency ratio & 4 \\
Action / proprioception dimension & 7 / 8 \\
Action and state normalization & Min--max \\
Validation fraction & 0 \\
Maximum text length & 128 tokens \\
\addlinespace[0.8em]
\multicolumn{2}{@{}l}{\textbf{Model and flow matching}} \\
\addlinespace[0.25em]
Backbone model & \texttt{Wan-AI/Wan2.2-TI2V-5B} \\
Tokenizer model & \texttt{Wan-AI/Wan2.1-T2V-1.3B} \\
Video--action axis alignment & Retracted 1D-RoPE \\
KV fusion & Lite KV-Fusion \\
Video layers / width / FFN width & 30 / 3,072 / 14,336 \\
Action layers / width / FFN width & 1 / 1,024 / 4,096 \\
Heads / head dimension (both streams) & 24 / 128 \\
Video patch / input and output channels & $(1,2,2)$ / 48 \\
Video attention mask & First-frame causal \\
Video action conditioning / action text attention & Disabled / disabled \\
Training / inference shift (both streams) & 5.0 / 5.0 \\
Training timesteps (both streams) & 1,000 \\
Video / action loss weight & 1.0 / 1.0 \\
\bottomrule
\end{tabularx}
\end{table}
\clearpage
\subsection{RoboTwin}
\label{app:reproducibility_robotwin}

\begin{table}[!htbp]
\centering
\footnotesize
\setlength{\tabcolsep}{4pt}
\renewcommand{\arraystretch}{1.05}
\caption{Faster-WAM training settings for RoboTwin.}
\label{tab:fasterwam_robotwin_reproducibility}
\begin{tabularx}{\textwidth}{@{}l>{\raggedright\arraybackslash}X@{}}
\toprule Setting & Value \\
\midrule
\multicolumn{2}{@{}l}{\textbf{Training}} \\
\addlinespace[0.25em]
Epochs / optimization steps & 5 / 58,710 \\
Global batch size & 512 \\
Batch per GPU / gradient accumulation & 16 / 2 \\
Optimizer / betas & AdamW / $(0.9,\,0.95)$ \\
Peak / final learning rate & $10^{-4}$ / $10^{-6}$ \\
Schedule / linear warmup & Cosine / 5\% of training steps \\
Weight decay / gradient clipping & 0.01 / 1.0 \\
Precision / random seed & BF16 / 42 \\
Video backbone & Jointly fine-tuned \\
\addlinespace[0.8em]
\multicolumn{2}{@{}l}{\textbf{Data and preprocessing}} \\
\addlinespace[0.25em]
Training data & RoboTwin 2.0 \\
Camera views & External, left wrist, and right wrist \\
Final image size per camera & Head: $256\times320$; each wrist: $128\times160$ \\
Model input image size & $384\times320$ \\
Sequence length / sampling stride & 33 / 1 \\
Action-to-video frequency ratio & 4 \\
Action / proprioception dimension & 14 / 14 \\
Action and state normalization & Z-score \\
Validation fraction & 0.01 \\
Maximum text length & 128 tokens \\
\addlinespace[0.8em]
\multicolumn{2}{@{}l}{\textbf{Model and flow matching}} \\
\addlinespace[0.25em]
Backbone model & \texttt{Wan-AI/Wan2.2-TI2V-5B} \\
Tokenizer model & \texttt{Wan-AI/Wan2.1-T2V-1.3B} \\
Video--action axis alignment & Retracted 1D-RoPE \\
KV fusion & Lite KV-Fusion \\
Video layers / width / FFN width & 30 / 3,072 / 14,336 \\
Action layers / width / FFN width & 1 / 1,024 / 4,096 \\
Heads / head dimension (both streams) & 24 / 128 \\
Video patch / input and output channels & $(1,2,2)$ / 48 \\
Video attention mask & First-frame causal \\
Video action conditioning / action text attention & Disabled / disabled \\
Training / inference shift (both streams) & 5.0 / 5.0 \\
Training timesteps (both streams) & 1,000 \\
Video / action loss weight & 1.0 / 1.0 \\
\bottomrule
\end{tabularx}
\end{table}

\clearpage

\end{document}